\documentclass[10pt,twocolumn,letterpaper]{article}

\usepackage{cvpr}              

\usepackage[accsupp]{axessibility}
\usepackage{times}
\usepackage{epsfig}
\usepackage{epigraph}
\usepackage{graphicx}
\usepackage{amsmath}
\usepackage{amssymb}
\usepackage[hypcap=false]{caption}
\usepackage[table]{xcolor}
\usepackage{tabularx}
\usepackage{multirow}
\usepackage{float}
\usepackage[pagebackref]{hyperref}
\hypersetup{colorlinks,linkcolor={red},citecolor={green},urlcolor={red}} 

\newcommand{\hquad}{\hspace{0.5em}} 

\DeclareMathOperator*{\minimize}{minimize \quad}

\definecolor{best_color}{HTML}{FCE5CD}
\definecolor{better_color}{HTML}{DEEDF2}

\newcommand{\pt}{\mathbf{x}} 
\newcommand{\cnlpt}{\pt_c}
\newcommand{\obspt}{\pt_o}
\newcommand{\ray}{\mathbf{r}} 
\newcommand{\cam}{\mathbf{e}} 

\newcommand{\nroffsetpack}{\Delta \pt}
\newcommand{\loss}{\mathcal{L}}

\newcommand{\cnlvolfunc}{F_c}
\newcommand{\obsvolfunc}{F_o} 
\newcommand{\posecorrectfunc}{P_{\rm pose}}

\newcommand{\motionfield}{T}
\newcommand{\skelmotionfield}{T_{\rm skel}}
\newcommand{\nrmotionfield}{T_{\rm NR}}

\newcommand{\rotbasis}{R}
\newcommand{\transbasis}{\mathbf{t}}

\newcommand{\weightcnl}{w_c}
\newcommand{\weightvolcnl}{W_c}
\newcommand{\weightobs}{w_o}

\newcommand{\wightvoldelta}{\Delta \weightvolcnl}
\newcommand{\weightvolgaussian}{W_g}

\newcommand{\weightcnn}{\rm CNN} 
\newcommand{\weightcnnlatent}{\textbf{z}}

\newcommand{\mlp}{\rm MLP} 
\newcommand{\posencode}{\gamma} 
\newcommand{\mlpcolor}{\mathbf{c}}
\newcommand{\mlpdensity}{\mathbf{\sigma}}

\newcommand{\bodypose}{\mathit{\mathbf{p}}}
\newcommand{\joints}{J}
\newcommand{\joint}{j}
\newcommand{\jangle}{\boldsymbol{\omega}}
\newcommand{\jangles}{\Omega}
\newcommand{\spacetfm}{M}

\newcommand{\volumerender}{\Gamma} 
\newcommand{\fglikelihood}{f}

\newcommand{\skelparam}{\theta_{\text{skel}}}
\newcommand{\nrparam}{\theta_{\rm NR}} 
\newcommand{\appearanceparam}{\theta_c} 
\newcommand{\posecorrectparam}{\theta_{\rm pose}}

\newcommand{\allparam}{\Theta}
\newcommand{\allparamdetail}{\appearanceparam, \skelparam, \nrparam, \posecorrectparam}

\begin{document}

\title{HumanNeRF:\\Free-viewpoint Rendering of Moving People from Monocular Video}
\author{Chung-Yi Weng$^1$ \hquad Brian Curless$^{1,2}$ \hquad Pratul P. Srinivasan$^{2}$ \hquad Jonathan T. Barron$^{2}$ \hquad Ira Kemelmacher-Shlizerman$^{1,2}$ \\ \\
$^1$University of Washington \quad $^2$Google Research
}

\twocolumn[{
\renewcommand\twocolumn[1][]{#1}
\maketitle
    \vspace*{-7ex}
    \begin{center}
    \includegraphics[width=\textwidth]{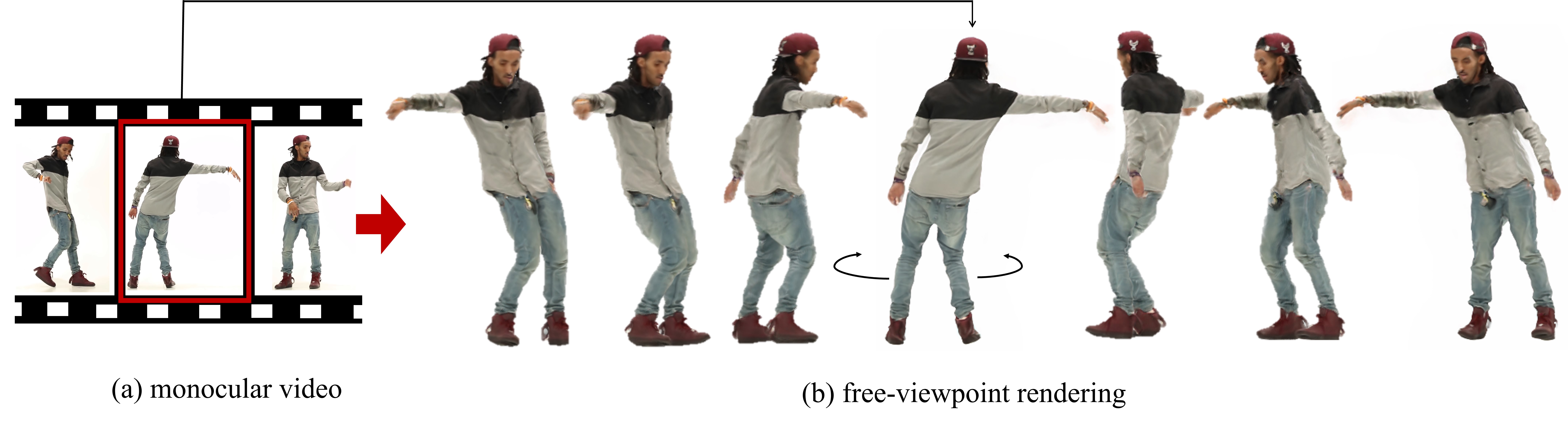}
    \end{center}  
    \vspace*{-20px}
    \captionof{figure}{Our method takes as input a monocular video{\protect\footnotemark} of a human performing complex movement, e.g., dancing (left), and creates a free-viewpoint rendering for any frame in the sequence (right). We construct a canonical subject appearance volume, and a motion field mapping from observation to canonical space, trained on the video. At test time, we take just the pose from the source frame (red square) and synthesize all output views, including the target view.  Please refer to the project page{\protect\footnotemark} to see the animated results.}
    \label{fig:teaser}
}
\vspace{10px}
]

\footnotetext[1]{e.g., {\scriptsize \url{https://youtu.be/0ORaAnJYROg}}}
\footnotetext[2]{\scriptsize {Project page: \url{https://grail.cs.washington.edu/projects/humannerf/}}}

\begin{abstract}
We introduce a free-viewpoint rendering method -- HumanNeRF -- that works on a given monocular video of a human performing complex body motions, e.g. a video from YouTube. Our method enables pausing the video at any frame and rendering the subject from arbitrary new camera viewpoints or even a full 360-degree camera path for that particular frame and body pose. This task is particularly challenging, as it requires synthesizing photorealistic details of the body, as seen from various camera angles that may not exist in the input video, as well as synthesizing fine details such as cloth folds and facial appearance. Our method optimizes for a volumetric representation of the person in a canonical T-pose, in concert with a motion field that maps the estimated canonical representation to every frame of the video via backward warps. The motion field is decomposed into skeletal rigid and non-rigid motions, produced by deep networks. We show significant performance improvements over prior work, and compelling examples of free-viewpoint renderings from monocular video of moving humans in challenging uncontrolled capture scenarios. 
\end{abstract}


\section{Introduction}

Given a single video of a human performing an activity, e.g., a YouTube or TikTok video of a dancer, we would like the ability to pause at any frame and rotate 360 degrees around the performer to view them from any angle at that moment in time (Figure 1). This problem -- free-viwepoint rendering of a moving subject -- is a longstanding research challenge, as it involves synthesizing previously unseen camera views while accounting for cloth folds, hair movement, and complex body poses \cite{kanade1997virtualized, matusik2000image, carranza2003free, starck2005video, vlasic2008articulated, guo2019relightables, casas20144d, flagg2009human}. The problem is particularly hard for the case of ``in-the-wild” videos taken with a single camera (monocular video), the case we address in this paper.  

Previous neural rendering methods \cite{martin2018lookingood, liu2020NeuralHumanRendering, peng2021neural, zhang2020nerf++, martinbrualla2020nerfw, wang2021ibrnet, barron2021mipnerf} typically assume multi-view input, careful lab capture, or do not perform well on humans due to non-rigid body motion. Human-specific methods typically assume a SMPL template~\cite{loper2015smpl} as a prior, which helps constrain the motion space but also introduces artifacts in clothing and complex motions that are not captured by the SMPL model \cite{peng2021neural, peng2021animatable}. Recently deformable NeRF methods~\cite{park2021nerfies, park2021hypernerf, tretschk2021nonrigid, pumarola2020d} perform well for small deformations, but not for large, full body motions like dancing. 

We  introduce a method, called HumanNeRF, that takes as input a single video of a moving person and, after per-frame, off-the-shelf segmentation (with some manual clean-up) and automatic 3D pose estimation, optimizes for a canonical, volumetric T-pose of the human together with motion field that maps the estimated canonical volume to each video frame via a backward warping. The motion field combines skeletal rigid motion with non-rigid motion, each represented volumetrically. Our solution is data-driven, with the canonical volume and motion fields derived from the video itself and optimized for large body deformations, trained end-to-end, including 3D pose refinement, without template models. At test time, we can pause at any frame in the video and, conditioned on the pose in that frame, render the resulting volumetric representation from any viewpoint.

We show results on a variety of examples: existing lab datasets, videos we captured outside the lab, and downloads from YouTube (with creator permission). Our method outperforms the state-of-the-art numerically and produces significantly higher visual quality. Please refer to the project page to see the results in motion.

\section{Related Work}

The physics of free-viewpoint rendering involves modeling geometry and surface properties and then rendering from new camera views. However, it remains difficult to recreate complex geometry and subtle lighting effects. Alternatively, image-based rendering \cite{shum2000imagerendering, szeliski2010computer} offers to render novel views based on given set of views in the image domain with a large corpus of research over the last couple decades \cite{levoy1996light, gortler1996lumigraph, debevec1996modeling, hedman2016scalable, hedman2018instant, chaurasia2013depth, chen1993view, zitnick2004high}. 

{\bf Human specific rendering:} The work of Kanade et al. \cite{kanade1997virtualized} is one of the earliest investigations into free-viewpoint rendering of humans. It introduced a dome equipped with cameras to recover depth maps and meshes, enabling novel views to be rendered by reprojecting and blending different views to account for mesh holes due to occlusions. Later, Matusik et al. \cite{matusik2000image} reconstructed a \textit{visual hull} from silhouettes of the subject and rendered it by carefully selecting pixels without an auxiliary geometric representation. Carranza et al. \cite{carranza2003free} used a parameterized body model as a prior and combined marker-less motion capture and view-dependent texturing \cite{debevec1996modeling}. Follow-on work introduced non-rigid deformation \cite{vlasic2008articulated}, texture warping \cite{xu2011video, casas20144d}, and various representations based on volumes \cite{de2008performance} or spheres \cite{starck2005video}. Collet et al. \cite{collet2015high} and Guo et al. \cite{guo2019relightables} build a system as well as pipeline that produces high-quality streamable \cite{collet2015high} or even relightable \cite{guo2019relightables} free-viewpoint videos of moving people.  

Most of these methods rely on multi-view videos -- typically expensive studio setups -- while we are interested in a simple monocular camera configuration.  

{\bf Neural radiance fields:} NeRF \cite{mildenhall2020nerf} and its extensions \cite{zhang2020nerf++, barron2021mipnerf, nerfactor, niemeyer2021giraffe, hedman2021baking, srinivasan2021nerv, tancik2020fourfeat} enable high quality rendering of novel views of static scenes. NeRF has recently been extended to dynamic scenes \cite{park2021nerfies, park2021hypernerf, tretschk2021nonrigid, Gao-freeviewvideo, li2021neural, pumarola2020d, xian2021space}, though these approaches generally assume that motion is small. 
We compare our method to these dynamic and deformable NeRF works in our results section.

{\bf Human-specific neural rendering:} The work of Liu et al. \cite{liu2020NeuralHumanRendering} starts from a pre-captured body model and learns to model time-dependent dynamic textures and enforce temporal coherence. Martin-Brualla et al. \cite{martin2018lookingood} trained a UNet to improve the artifacts introduced by volumetric capture. The follow-up work of Pandey et al. \cite{pandey2019volumetric} reduced the number of required input frames to as few as a single RGBD image via semi-parametric learning.  Wu et al. \cite{wu2020multi} and Peng et al. \cite{peng2021neural} explored the use of learned structured latent codes embedded for point clouds (from MVS \cite{schoenberger2016mvs}) or reposed mesh vertices (from SMPL \cite{loper2015smpl}) and learn an accompanying UNet- or NeRF-based neural renderer. Zhang et al. \cite{zhang2021stnerf} decomposed a scene into background and individual performers, and represented them with separated NeRFs thus enabling scene editing. Other than free-viewpoint rendering, there is another related active research field that focus on human motion retargeting either in 2D \cite{wang2018video, chan2019everybody, sarkar2021style, balakrishnan2018synthesizing, neverova2018dense, ma2017pose, wang2021dance} or 3D \cite{weng2020vid2actor, liu2021neural, peng2021animatable, habermann2021, sanyal2021learning,  yang2021s3, huang2020arch, he2021arch++}. The main difference between our method and those works is that we take as input \textit{monocular} video that contains \textit{complex} human motions and enable high-fidelity full 3D rendering.

Additionally, our formulation of skeletal motion draws inspiration from Vid2Actor proposed by Weng et al. \cite{weng2020vid2actor}, a method intended for rigidly animatable characters. Instead, we focus on the free-viewpoint application and recovering pose-dependent, non-rigid deformation and outperform them significantly for this application.

\textbf{Concurrent work:}  Xu et al. \cite{xu2021h} co-learn implicit geometry as well as appearance from images. They largely focus on multi-view setups with a few examples on monocular videos where the human motion is simple (A-pose). Su et al. \cite{su2021anerf} use an over-parameterized NeRF to rigidly transform NeRF features for refining body pose and thus final rendering. The non-rigid motion is not explicitly modeled and the rendering quality is not high. A similar approach is discovered by Noguchi et al. \cite{2021narf} as well but still shows results of limited visual quality.

\section{Representing a Human as a Neural Field}

\begin{figure*}
  \centering
  \includegraphics[width=\textwidth]{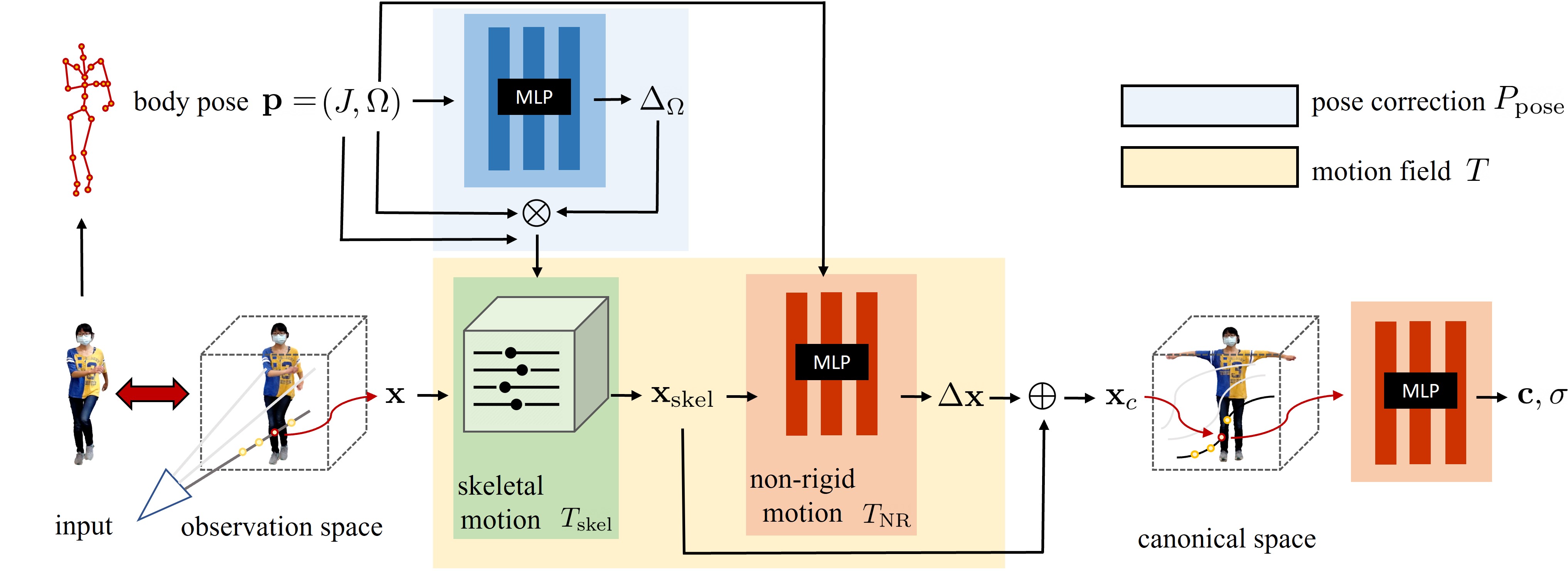}
  \vspace{-14px}
  \caption{Our method takes a video frame as input and optimizes for canonical appearance, represented as a continuous field, as well as a motion field mapping from observation to canonical space. The motion field is decomposed into skeletal rigid and non-rigid motion, represented as a discrete grid and a continuous field respectively. We additionally refine body pose initialized with an off-the-shelf body pose estimator, leading to better alignment. A loss is imposed between the volume rendering in observation space and the input image, directing optimization towards a solution.} 
  \label{fig:overview}
  \vspace{-14px}
\end{figure*}

We represent a moving person with a canonical appearance volume $\cnlvolfunc$ warped to an observed pose to produce output appearance volume $F_o$:
\begin{equation}
\label{eq:canonical_observation_mapping}
\obsvolfunc(\pt, \bodypose) = \cnlvolfunc(\motionfield(\pt,\bodypose)),
\end{equation}
where $\cnlvolfunc: \pt \rightarrow (\mlpcolor, \mlpdensity)$ maps position $\pt$ to color $\mlpcolor$ and density $\mlpdensity$, and $\motionfield: (\obspt, \bodypose) \rightarrow \cnlpt$ defines a motion field mapping points from observed space back to canonical space, guided by observed pose $\bodypose = (\joints, \jangles)$, where $\joints$ includes $K$ standard 3D joint locations, and $\jangles = \{\jangle_i$\} are local joint rotations represented as axis-angle vectors $\jangle_i$.

We handle complex human movement with complex deformation by decomposing the motion field into two parts:
\begin{equation}
\label{eq:motion_factoring}
\motionfield(\pt,\bodypose) = \skelmotionfield(\pt,\bodypose) 
  + \nrmotionfield(\skelmotionfield(\pt,\bodypose), \bodypose),
\end{equation}
where $\skelmotionfield$ represents skeleton-driven deformation, essentially inverse (volumetric) linear-blend skinning, and $\nrmotionfield$ starts from the skeleton-driven deformation and produces an offset $\nroffsetpack$ to it.  In effect, $\skelmotionfield$ provides the coarse deformation driven by standard skinning, and $\nrmotionfield$ provides the more non-rigid effects, e.g., due to deformation of clothing.

For ``in-the-wild'' imagery, we use an off-the-shelf 3D body+camera pose estimator.  Due to inaccuracy in pose estimation, we also solve for a pose correction function $\posecorrectfunc(\bodypose)$ that better explains the observations, and apply this improvement to the skeleton-driven deformation, i.e., we replace $\skelmotionfield(\pt,\bodypose)$ with $\skelmotionfield(\pt,\posecorrectfunc(\bodypose))$ in Eq. \ref{eq:motion_factoring}.

Figure \ref{fig:overview} gives an overview of the components of our system.  In the following sections, we describe these components in detail.

{\bf Canonical volume:}  We represent the canonical volume $\cnlvolfunc$ as a continuous field with an MLP that outputs color $\mlpcolor$ and density $\mlpdensity$ given a point $\pt$:
 
\begin{equation}
\cnlvolfunc(\pt) = {\rm MLP}_{\appearanceparam}(\posencode(\pt)), 
\end{equation}
where $\posencode$ is a sinusoidal positional encoding defined as $(\pt, \sin(2^{0}\pi\pt), \cos(2^{0}\pi\pt), ..., \sin(2^{L-1}\pi\pt), \cos(2^{L-1}\pi\pt))$ and $L$ is a hyper-parameter that determines the number of frequency bands \cite{mildenhall2020nerf}. 

{\bf Skeletal motion:} Following Weng et al. \cite{weng2020vid2actor}, we compute the skeletal deformation $\skelmotionfield$ as a kind of inverse, linear blend skinning that maps points in the observation space to the canonical space:
\begin{equation}
T_{\rm skel}(\pt, \bodypose) = \
  \sum_{i=1}^{K}{\weightobs^i(\pt){(\rotbasis_{i}\pt + \transbasis_{i})}},
\label{eq:skel_motion}  
\end{equation}
where $\weightobs^i$ is the blend weight for the $i$-th bone and $\rotbasis_i$, $\transbasis_i$ are the rotation and translation, respectively, that map the bone's coordinates from observation to canonical space; $\rotbasis_i$ and $\transbasis_i$ can be explicitly computed from $\bodypose$ (see supplementary).  We then aim to optimize for $\weightobs^i$. 

In practice, we solve for $\weightcnl^i$ defined in canonical space by storing $K$ blend weights as a set of volumes $\{\weightcnl^i(\pt)\}$, from which the observation weights are derived as:
\begin{equation}
    \weightobs^i(\pt) =  \frac{\weightcnl^i(\rotbasis_{i}\pt + \transbasis_{i})}
  {\sum_{k=1}^{K}{\weightcnl^k(\rotbasis_{k}\pt + \transbasis_{k})}}.
\label{eq:skel_motion_skin}  
\end{equation}

Solving for a single set of weight volumes $\{\weightcnl^i(\pt)\}$ in canonical space, instead of $N$ sets of $\{\weightobs^i(\pt)\}$ in observation space (corresponding to $N$ input images),  can lead to better generalization as it avoids over-fitting  \cite{weng2020vid2actor, chen2021snarf}.

We pack the set of $\{\weightcnl^i(\pt)\}$ into a single volume $\weightvolcnl(\pt)$ with $K$ channels. Rather than encode $\weightvolcnl$ with an MLP, we choose an explicit volume representation for two reasons: (1)~Eq.~\ref{eq:skel_motion_skin} shows that $K$ MLP evaluations  would be needed to compute each $\weightobs^i(\pt)$, infeasible for optimization ($K=24$ in our work); (2) an explicit volume with limited resolution resampled via trilinear interpolation provides smoothness that can help regularize the optimization later. In practice, during optimization, rather than directly solve for volume $\weightvolcnl$, we solve for parameters $\skelparam$ of a $\weightcnn$ that generates the volume from a random (constant) latent code $\weightcnnlatent$: 
\begin{equation}
\weightvolcnl(\pt) = \weightcnn_{\skelparam}(\pt; \weightcnnlatent).
\end{equation}

We also add one more channel, a background class, and represent $\weightvolcnl$ as a volume with $K+1$ channels. We then apply channel-wise \textit{softmax} to the output of the $\weightcnn$, enforcing a partition of unity across the channels. The denominator of Eq.~\ref{eq:skel_motion_skin} can then be used to approximate likelihood $\fglikelihood(\pt)$ of being part of the subject, where $\fglikelihood(\pt) = \sum_{k=1}^{K}{\weightcnl^k(\rotbasis_{k}\pt + \transbasis_{k})}$.  When $\fglikelihood(\pt)$ is close to zero, we are likely in free space away from the subject, which we will use during volume rendering.  

The idea of optimizing blend weights (or skinning field) is not new. Similar approaches have been applied to human modeling~\cite{deng2019neural, huang2020arch, chen2021snarf, LEAP:CVPR:21, Saito:CVPR:2021, yang2021s3, peng2021animatable, tiwari2021neural, bhatnagar2020loopreg}. Our formulation follows Weng et al.~\cite{weng2020vid2actor}, but also shares similarities with Tiwari et al.~\cite{tiwari2021neural}; the latter learns from 3D scans while we learn from 2D images, like the former.

{\bf Non-rigid motion:} We represent non-rigid motion $\nrmotionfield$ as an offset $\nroffsetpack$ to the skeleton-driven motion, conditioned on that motion, i.e., $\nroffsetpack(\pt, \bodypose) = \nrmotionfield(\skelmotionfield(\pt,\bodypose),\bodypose))$.  To capture detail, we represent $T_{\rm NR}$ with an MLP:
\begin{equation}
\nrmotionfield(\pt, \bodypose) 
  = {\mlp}_{\nrparam}(\posencode(\pt); \jangles),
\end{equation}
where again we use the standard positional encoding  $\posencode$ and condition the $\mlp$ on $\jangles$, the joint angles of body pose $\bodypose$.

{\bf Pose correction:} The body pose $\bodypose = (\joints,\jangles)$ estimated from an image is often inaccurate.  To address this, we solve for an update to the pose:
\begin{equation}
\posecorrectfunc(\bodypose) = (\joints, \Delta_\jangles(\bodypose) \otimes \jangles),
\end{equation}
where we hold the joints $\joints$ fixed and optimize for a relative update to the joint angles, $\Delta_\jangles = (\Delta \jangle_0, ..., \Delta \jangle_K)$ which is then applied to $\jangles$ to get updated rotation vectors.

Empirically we found, instead of directly optimizing for $\Delta_\jangles$, solving for the parameters $\posecorrectparam$ of an $\mlp$ that generates $\Delta_\jangles$ conditioned on $\jangles$ leads to faster convergence:
\begin{equation}
\Delta_\jangles(\bodypose) = {\mlp}_{\posecorrectparam}(\jangles).
\end{equation}

With this pose correction, we can re-write the equation that warps from observation space to canonical space as:
\begin{equation}
\motionfield(\pt,\bodypose) = \skelmotionfield(\pt,\posecorrectfunc(\bodypose)) 
  + \nrmotionfield(\skelmotionfield(\pt,\posecorrectfunc(\bodypose)), \bodypose)
\end{equation}

\section{Optimizing a HumanNeRF}

\renewcommand{\arraystretch}{1.2}
\begin{table*}[htbp]
\centering
\begin{tabular}{|c || c | c | c || c | c | c || c | c | c|}
\hline
\multirow{2}{*}{} &  \multicolumn{3}{c||}{Subject \textbf{377}} & \multicolumn{3}{c||}{Subject \textbf{386}} & \multicolumn{3}{c|}{Subject \textbf{387}} \\ 
\cline{2-10}
 & PSNR $\uparrow$ & SSIM $\uparrow$ & LPIPS* $\downarrow$ & PSNR $\uparrow$ & SSIM $\uparrow$ & LPIPS* $\downarrow $ & PSNR $\uparrow$ & SSIM $\uparrow$ & LPIPS* $\downarrow$ \\ 
\hline
Neural Body \cite{peng2021neural} & 29.11 & 0.9674 & 40.95 & 30.54 & 0.9678 & 46.43 & 27.00 & 0.9518 & 59.47 \\
\hline
Ours & \cellcolor{best_color}30.41  & \cellcolor{best_color}0.9743 & \cellcolor{best_color}24.06 & \cellcolor{best_color}33.20 & \cellcolor{best_color}0.9752 & \cellcolor{best_color}28.99 & \cellcolor{best_color}28.18 & \cellcolor{best_color}0.9632 & \cellcolor{best_color}35.58 \\
\hline
\hline
\multirow{2}{*}{} &  \multicolumn{3}{c||}{Subject \textbf{392}} & \multicolumn{3}{c||}{Subject \textbf{393}} & \multicolumn{3}{c|}{Subject \textbf{394}} \\ 
\cline{2-10}
 & PSNR $\uparrow$ & SSIM $\uparrow$ & LPIPS* $\downarrow$ & PSNR $\uparrow$ & SSIM $\uparrow$ & LPIPS* $\downarrow $ & PSNR $\uparrow$ & SSIM $\uparrow$ & LPIPS* $\downarrow$ \\ 
\hline
Neural Body\cite{peng2021neural} & 30.10 & 0.9642 & 53.27 & \cellcolor{best_color}28.61 & 0.9590 & 59.05 & 29.10 & 0.9593 & 54.55\\
\hline
Ours & \cellcolor{best_color}31.04 & \cellcolor{best_color}0.9705 & \cellcolor{best_color}32.12 & 28.31 & \cellcolor{best_color}0.9603 & \cellcolor{best_color}36.72 & \cellcolor{best_color}30.31 & \cellcolor{best_color}0.9642 & \cellcolor{best_color}32.89\\
\hline
\end{tabular}
\caption{Quantitative comparison on ZJU-MoCap dataset. We color cells that have the \colorbox{best_color}{best} metric value. LPIPS* = LPIPS $\times 10^3$.}
\label{table:zju_number_vs_nb}
\vspace{-14px}
\end{table*}


In this section, we describe the overall objective function we minimize, our volume rendering procedure,  how we regularize the optimization process, specific loss function details, and the ray sampling method.

{\bf HumanNeRF objective:} Given input frames $\{I_1, I_2, ..., I_N\}$, body poses $\{\bodypose_1, \bodypose_2, ..., \bodypose_N\}$, and cameras $\{\cam_1, \cam_2, ..., \cam_N\}$, we are solving the problem:
\begin{equation}
\label{eq:optim_problem_def}
    \minimize_{\allparam} \sum_{i=1}^N \loss\{\volumerender[\cnlvolfunc(\motionfield(\pt, \bodypose_i)), \cam_i], I_i\},
\end{equation}
where $\loss\{\cdot\}$ is the loss function and $\volumerender[\cdot]$ is a volume renderer, and we minimize the loss with respect to all network parameters $\allparam = \{\allparamdetail\}$.  As we have seen, $\cnlvolfunc$ is determined by parameters $\appearanceparam$, while the transformation $T$ from observation space to canonical space relies on parameters $\skelparam$, $\nrparam$, and $\posecorrectparam$.

\subsection{Volume rendering}

We render a neural field using the volume rendering equation \cite{max1995optical} as described by Mildenhall et al. \cite{mildenhall2020nerf}. The expected color $C(\ray)$ of a ray $\ray$ with $D$ samples can be written as:
\begin{equation}
\label{eq:volume_rendering}
\begin{aligned}
    & C(\ray) = \sum_{i=1}^{D} (\prod_{j=1}^{i-1} (1 - \alpha_j))\alpha_i\mlpcolor(\pt_i), \\
    & \quad \alpha_i = 1 - \exp(-\mlpdensity(\pt_i) \Delta t_i), 
\end{aligned}
\end{equation}
where $\Delta t_i$ is the interval between sample $i$ and $i+1$. 

We further augment the definition of $\alpha_i$ to be small when approximate foreground probability $\fglikelihood(\pt)$ is low:

\begin{equation}
\label{eq:alpha_with_validity}
    \alpha_i = \fglikelihood(\pt_i)(1 - \exp(-\mlpdensity(\pt_i) \Delta t_i)), 
\end{equation}

We apply the stratified sampling approach proposed by NeRF~\cite{mildenhall2020nerf}. We do not use hierarchical sampling since the bounding box of a subject can be estimated from their 3D body pose. We then only sample points inside the box.

\subsection{Delayed optimization of non-rigid motion field}

When solving for all the network parameters in Eq. \ref{eq:optim_problem_def} at once, we find that the the optimized skeleton-driven and non-rigid motions are not decoupled -- a portion of the subject's skeletal motions is modeled by the non-rigid motion field -- due to over-fitting of non-rigid motions to the input images. As a result, the quality degrades when rendering unseen views. 
 
We manage the optimization process to solve the problem. Specifically, we disable non-rigid motions at the beginning of optimization, and then bring them back in a coarse-to-fine manner~\cite{park2021nerfies, hertz2021sape}. To achieve this, for the non-rigid motion MLP, we apply a truncated Hann window to its frequency bands of positional encoding, to prevent overfitting to the data \cite{tancik2020fourfeat}, increasing the window size as the optimization proceeds. Following Park et al. \cite{park2021nerfies}, we define the weight for each frequency band $j$ of positional encoding:
\begin{equation}
\label{eq:truncated_hann_window}
    \mathit{w}(\tau) = \frac{1 - \cos({\rm clamp}(\tau - j, 0, 1)\pi)}{2}, 
\end{equation}
where $\tau \in [0, L)$ determines the width of a truncated Hann window, and $L$ is the total number of frequency bands in positional encoding. We then define $\tau$ as a function of the optimization iteration:
\begin{equation}
\label{eq:alpha_define}
    \tau(t) = L\frac{\max(0, t - T_s)}{T_e - T_s},
\end{equation}
where $t$ is the current iteration, and $T_s$ and $T_e$ are hyper-parameters that determine when to enable non-rigid motion optimization and when to use full frequency bands of positional encoding. We remove position identity from positional encoding without affecting performance \cite{barron2021mipnerf}. By doing so, we can completely disable non-rigid motion optimization by setting $\tau=0$ \cite{park2021hypernerf}.

\subsection{Loss and ray sampling}

\textbf{Loss function:} We employ both an MSE loss to match pixel-wise appearance and a perceptual loss, LPIPS~\cite{zhang2018unreasonable}, to provide robustness to slight misalignments and shading variation and to improve detail in the reconstruction.  Our final loss function is $\loss = \loss_{\text{LPIPS}} + \lambda \loss_{\text{MSE}}$. We use $\lambda = 0.2$ and choose VGG as the backbone of LPIPS.

\textbf{Patch-based ray sampling:} Training on random ray samples, as done in NeRF \cite{mildenhall2020nerf}, does not work for minimizing our loss because LPIPS uses convolutions to extract features. Instead, we sample $G$ patches with size $H \times H$ on an image, and render a total of $G \times H \times H$ rays in each batch. The rendered patch is compared against the patch with the same position on the input image. We use $G=6$ and $H=32$ in our experiments. Similar approaches were also used in NeRF-based generative models \cite{Schwarz2020NEURIPS}.
\section{Results}

\subsection{Evaluation dataset} 

We evaluate our method on the ZJU-MoCap dataset \cite{peng2021neural}, self-captured data (\textit{rugby, hoodie}), and YouTube videos downloaded from Internet (\textit{story}\footnote{\url{https://youtu.be/0ORaAnJYROg}}, \textit{way2sexy}\footnote{\url{https://youtu.be/gEpJDE8ZbhU}}, \textit{invisible}\footnote{\url{https://youtu.be/ANwEiICt7BM}}). All subjects in these videos provided consent to use their data. For ZJU-MoCap, we select 6 subjects (377, 386, 387, 392, 393, 394) with diverse motions and use images captured by ``\textit{camera 1}'' as input and the other 22 cameras for evaluation. We directly apply camera matrices, body pose, and segmentation provided by the dataset. For videos ``in the wild'' (self-captured and YouTube videos), we run SPIN~\cite{kolotouros2019spin} to get approximate camera and body pose, automatically segment the foreground subject, and then manually correct errors in the segmentation.  (High quality segmentation is necessary for best results; purely automatic segmenters were not accurate enough, and improving on them was outside the scope of this paper, an area of future work.) We additionally resize video frames to keep the height of subject at approximately 500 pixels. 

\begin{figure*}
  \centering
  \includegraphics[width=0.95\textwidth]{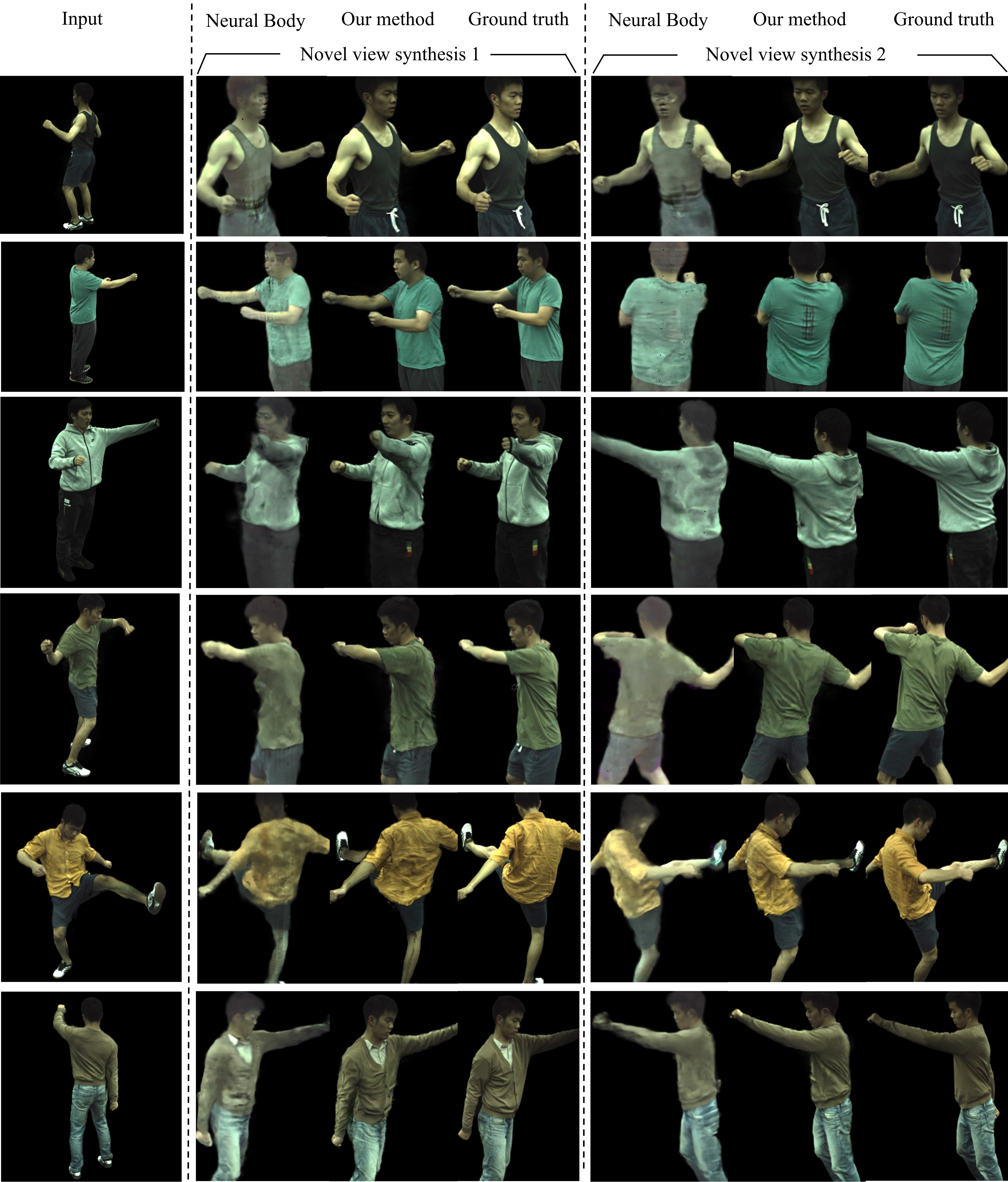}
  \caption{Qualitative comparison on ZJU-MoCap dataset.}
  \vspace{-10px}
  \label{fig:zju_mocap_vs_nb}
\end{figure*}

\subsection{Optimization details} 

We optimize Eq. \ref{eq:optim_problem_def} using the Adam optimizer \cite{kingma2014adam} with hyperparameters $\beta_1 = 0.9$ and $\beta_2 = 0.99$. We set the learning rate to $5 \times 10^{-4}$ for $\appearanceparam$ (the canonical $\mlp$), and $5 \times 10^{-5}$ for all the others. We use 128 samples per ray. The optimization takes $400K$ iterations (about 72 hours) on 4 GeForce RTX 2080 Ti GPUs. We apply delayed optimization with $T_s=10K$ and $T_e=50K$ to ZJU-MoCap data, and with $T_s=100K$ and $T_e=200K$ to the others. In addition, we postpone pose refinement until after 20K iterations for in-the-wild videos.

\subsection{Evaluation method} 

\vspace{-6px}
\begin{table}[H]
\centering
\newcolumntype{Y}[1]{>{\hsize=#1\hsize}X}%
\begin{tabularx}{\linewidth}{|Y{0.26}|Y{0.58}|Y{0.58}|Y{0.58}|}
\hline
{} & \small{Neural Body} & \small{HyperNeRF} & \small{HumanNeRF}\\
\hline
\small{Setup} & \small{multi-camera} & \small{ single camera} & \small{single camera}\\
\hline
\small{Subject} & \small{dynamic \newline human} & \small{quasi-static \newline general scene} & \small{dynamic \newline human}\\
\hline
\small{Priors} & \small{body pose, \newline SMPL vertices \newline (reposed)} & \small{rigidity} & \small{body pose \newline (approx.)} \\
\hline
\end{tabularx}
\caption{Differences between the compared methods.}
\vspace{-10px}
\label{table:method_difference}
\end{table}

We compare our method with Neural Body~\cite{peng2021neural} (typically used with multiple cameras) and HyperNeRF~\cite{park2021hypernerf} (single moving camera around the subject), state-of-the-art methods for modeling humans and general scenes for novel view synthesis. Our method works with a single camera which can be static or moving; we focus on results with a static camera and moving subjects, a natural way to capture a person's performance. The differences between these methods are listed in Table \ref{table:method_difference}. 

\subsection{Comparisons}

\begin{figure}
\centering
\includegraphics[width=\linewidth]{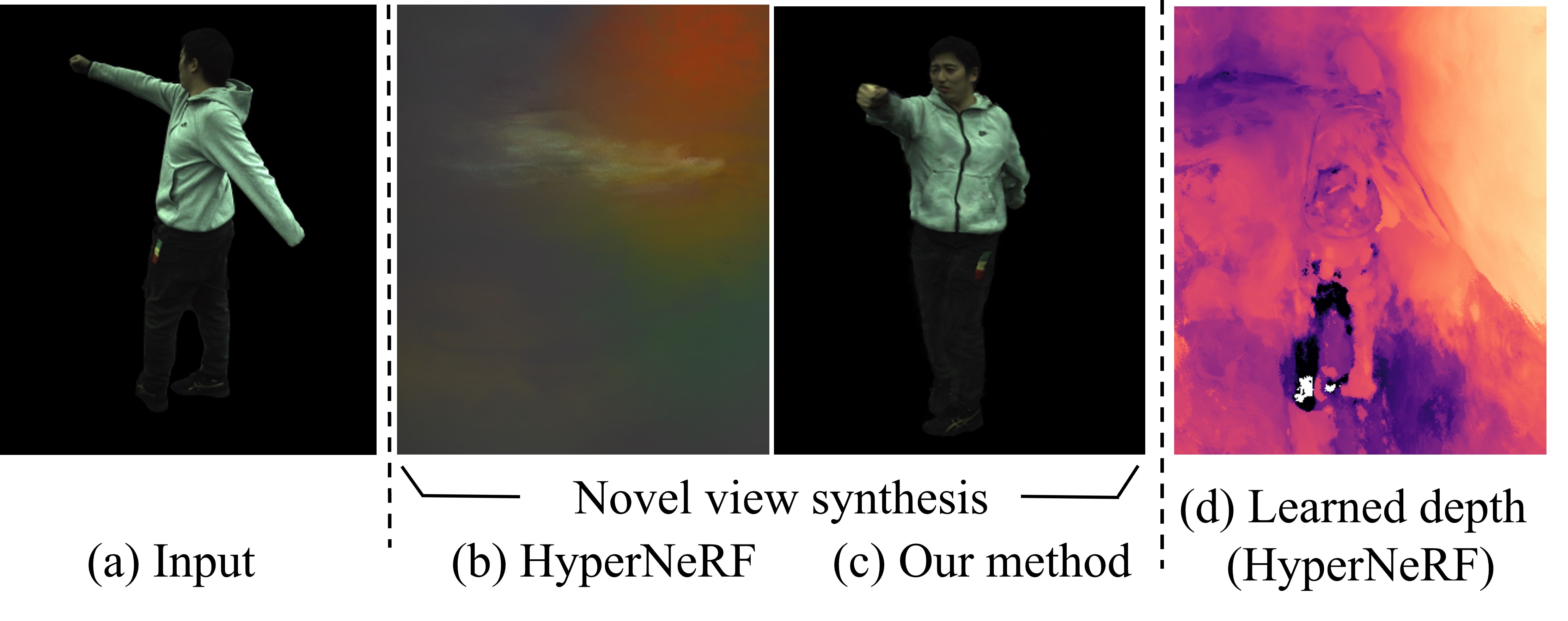}
\caption{Qualitative comparison to HyperNeRF \cite{park2021hypernerf}}
\label{fig:zju_mocap_vs_hypernerf}
\vspace{-10px}
\end{figure}

We found HyperNeRF does not produce meaningful output for novel view synthesis in our experiments, as shown in Fig. \ref{fig:zju_mocap_vs_hypernerf}, likely because it relies on multiple views (moving camera) to build a coherent 3D model.  For the static camera case with moving subject, it fails to recover a meaningful depth map and appears to memorizes the input images rather than generalize from them. We note that dynamic human motions are also more extreme than the examples shown to work with HyperNeRF.  

\begin{figure*}
\centering
\vspace{-8px}
\includegraphics[width=\textwidth]{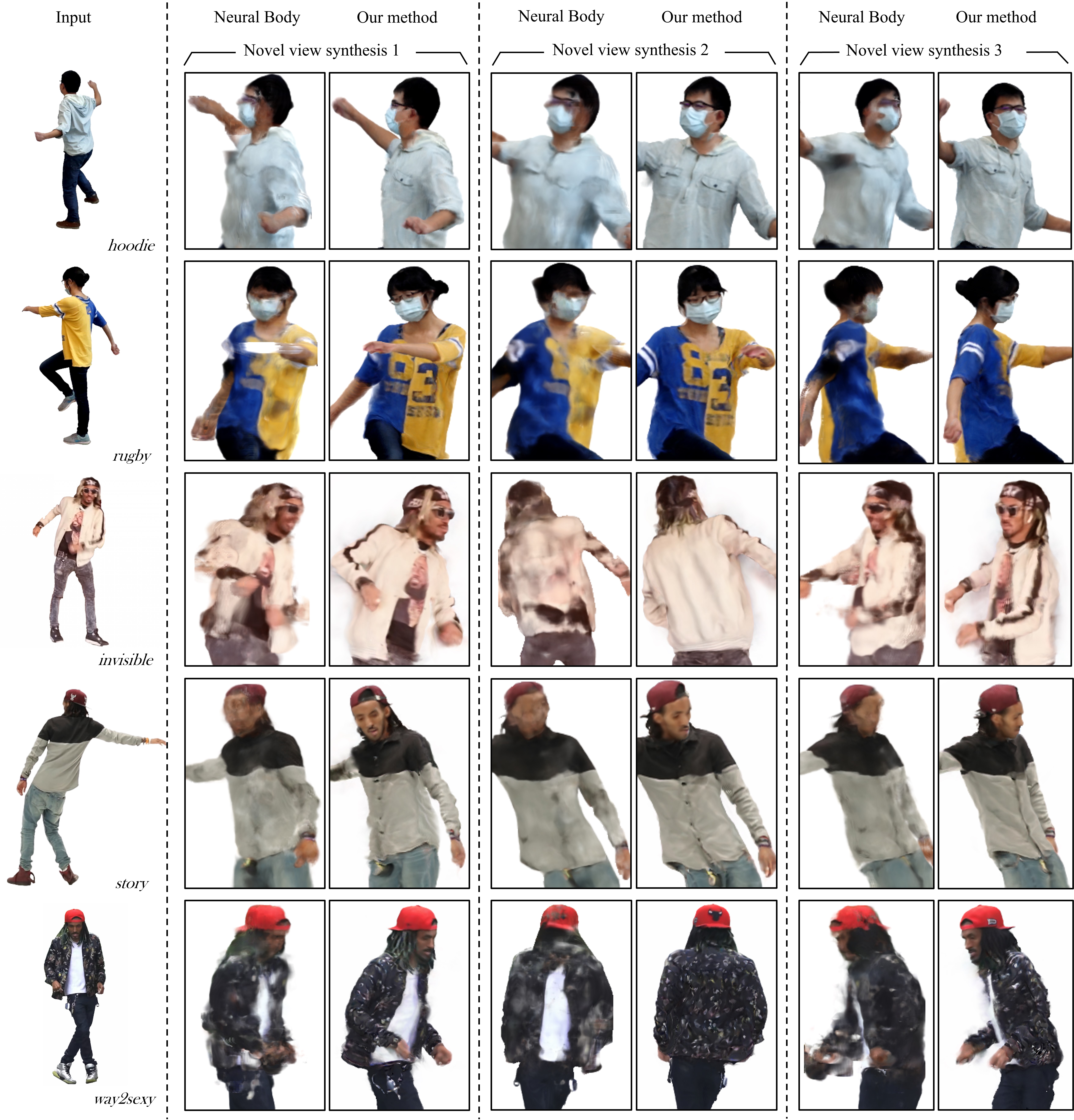}
\caption{Qualitative comparison for self-captured videos (first two rows) and YouTube videos (bottom three).}
\label{fig:wild_vs_nb}
\vspace{-16px}
\end{figure*}

Quantitatively, as shown in Table~\ref{table:zju_number_vs_nb}, HumanNeRF outperforms Neural Body for all subjects and under all metrics, except for subject 393 on PSNR (a metric known to favor smooth results \cite{zhang2018unreasonable}). The gain is particularly significant with perceptual metric LPIPS, nearly 40\% improvement on average. Fig. \ref{fig:zju_mocap_vs_nb} shows that HumanNeRF's visual quality is substantially better then Neural Body for this dataset.  Our method is capable of producing high fidelity details similar to the ground truth even on completely unobserved views, while Neural Body tends to produce blurrier results. The results for self-captured and YouTube videos, shown in Fig. ~\ref{fig:wild_vs_nb}, also show consistently higher quality reconstructions with HumanNeRF.

\subsection{Ablation studies}

\begin{figure}[H]
\centering
\includegraphics[width=\linewidth]{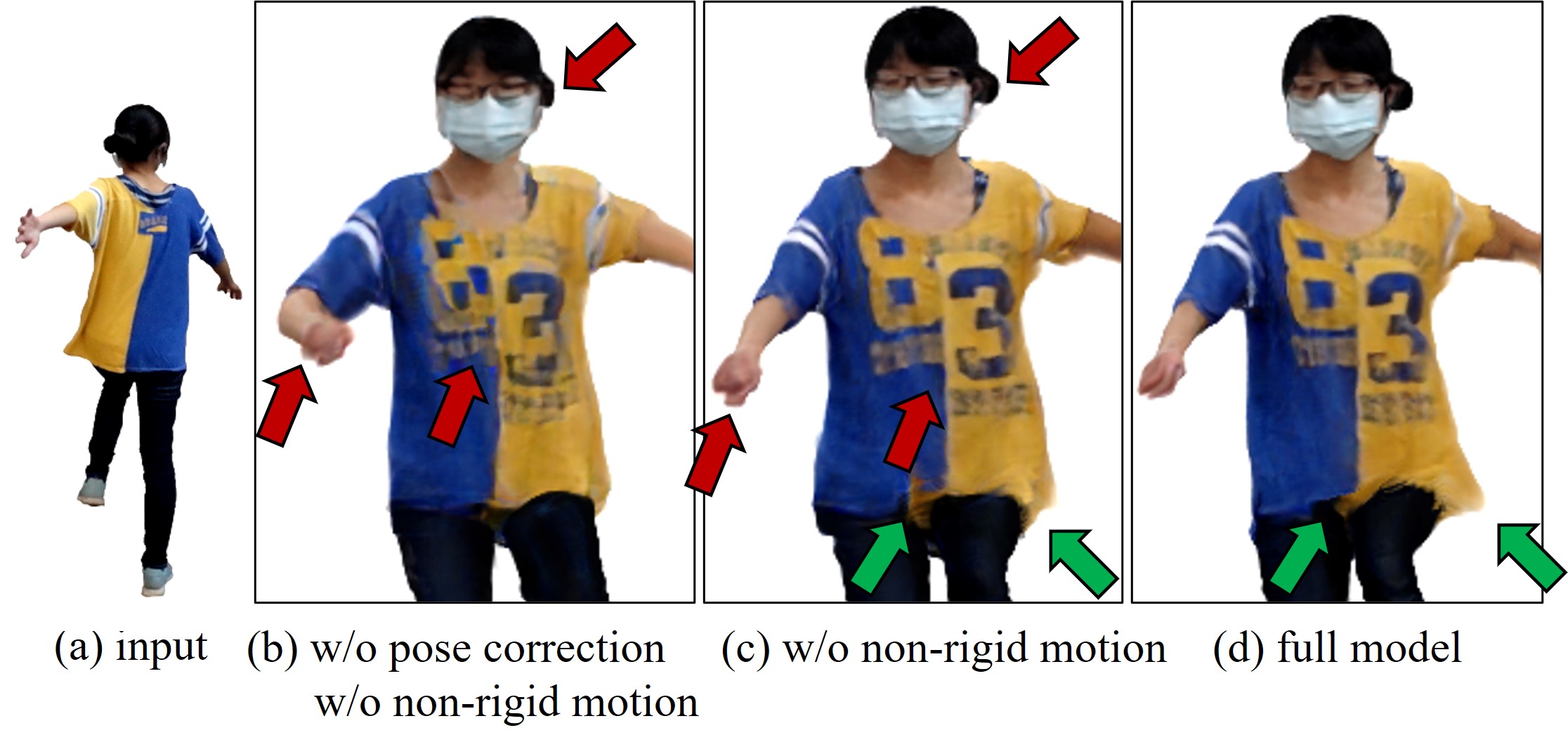}
\caption{Pose correction and non-rigid motion improve novel view synthesis. Pose correction straightens the right arm and adds details (red arrows in (b) vs (c)) and non-rigid deformation improves clothing alignment and shape (green arrows in (c) vs. (d)).}
\label{fig:ablation_pose_nr_motion_novel_view}
\vspace{-10px}
\end{figure}

Table 3 illustrates that skeletal deformation alone is enough for significant improvement over Neural Body for the ZJU-MoCap data.  Adding non-rigid deformation provides further gains.  (Accurate poses were provided for this dataset, thus we did not perform an ablation for the pose optimizer here.)

Fig.~\ref{fig:ablation_pose_nr_motion_novel_view} shows visually, for in-the-wild data, the importance of including non-rigid motion and, additionally, pose correction for an unseen view.

\begin{table}[H]
\centering
\begin{tabular}{|c|c|c|c|}
\hline
 & PSNR $\uparrow$ & SSIM $\uparrow$ & LPIPS* $\downarrow$ \\
\hline 
\hline
Neural Body \cite{peng2021neural}& 29.08 & 0.9616 & 52.27 \\
\hline
Ours (w/o non-rigid) & \cellcolor{better_color}29.81 & \cellcolor{better_color}0.9657 &  \cellcolor{better_color}34.17 \\
\hline
Ours (full model) & \cellcolor{best_color}30.24 & \cellcolor{best_color}0.9679 & \cellcolor{best_color}31.73 \\
\hline
\end{tabular}
\vspace{-6px}
\caption{Ablation study on ZJU-MoCap. We compute averages over 6 sequences. We color cells with best \colorbox{best_color}{best} and \colorbox{better_color}{second best} metric values. LPIPS* = LPIPS $\times 10^3$.}
\label{table:ablation_zju}
\vspace{-10px}
\end{table}

\begin{figure}[H]
\centering
\vspace{-10px}
\includegraphics[width=\linewidth]{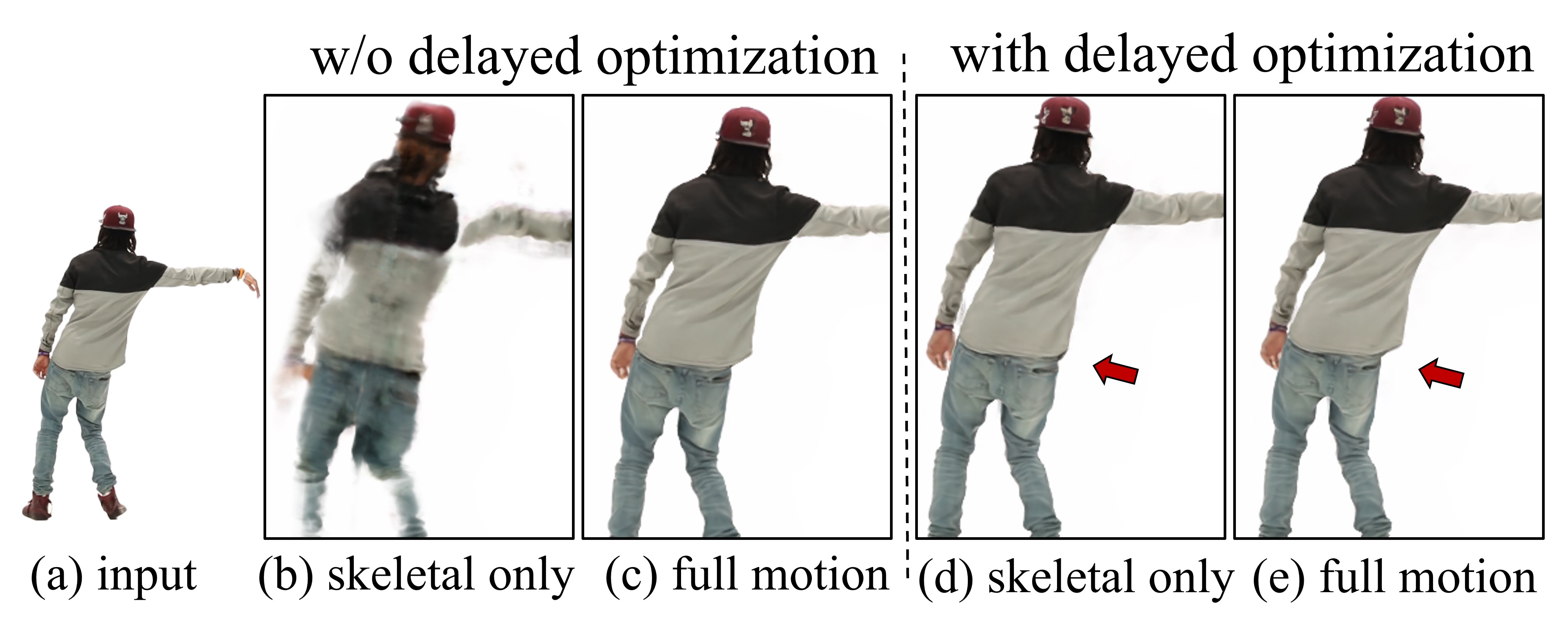}
\vspace{-20px}
\caption{Delayed optimization (d, e) leads to better motion decoupling than the result without it (b, c).  The skeletal-only deformation result without delayed optimization is poor, which can ``corrected'' by the non-rigid deformation, but leads to poor view generalization (below).}
\label{fig:delay_nr_motion_decouple}
\vspace{-8px}
\end{figure}

\begin{figure}[H]
\centering
\includegraphics[width=0.85\linewidth]{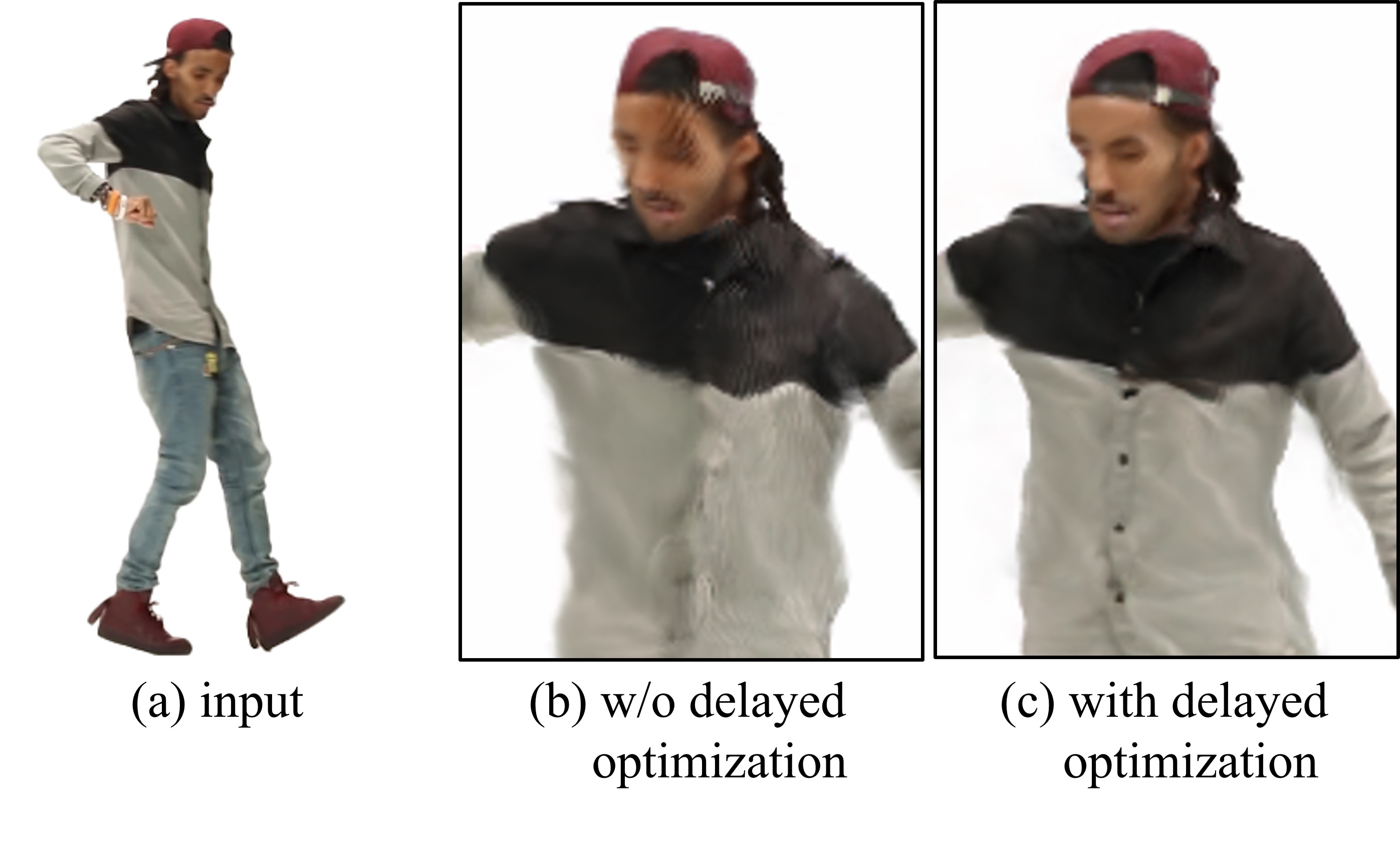}
\vspace{-16px}
\caption{Without delayed optimization and strong decoupling of skeletal and non-rigd deformations, generalization to unseen views is poor (b).  With delayed optimization, the decoupling leads to good generalization (c).}
\label{fig:delay_nr_motion_novel_view}
\vspace{-6px}
\end{figure}

Fig.~\ref{fig:delay_nr_motion_decouple} shows the importance of delayed optimization for decoupling skeletal deformation and non-rigid deformation.  When not decoupled well, generalization to new views is much poorer, as shown in Fig.~\ref{fig:delay_nr_motion_novel_view}.

\section{Discussion}

\vspace{-4px}
\textbf{Limitations:} Our method has artifacts when part of the body is not shown in the video. Pose correction improves image alignment but may fail if the initial pose estimate is poor or if the image contains strong artifacts such as motion blur. In addition, we observed the frame-by-frame body poses are still not temporally smooth even after pose correction. We assume non-rigid motion is pose-dependent, but this is not always true (e.g., clothes shifting due to wind or due to follow-through after dynamic subject motion). We also assume fairly diffuse lighting, so that appearance does not change dramatically as the points on the subject rotate around.  Finally, for in-the-wild videos, we rely on manual intervention to correct segmentation errors. These limitations point to a range of interesting avenues for future work.

\textbf{Conclusion:} We have presented HumanNeRF, producing state-of-the-art results for free-viewpoint renderings of moving people from monocular video. We demonstrate high fidelity results for this challenging scenario by carefully modeling body pose and motion as well as regularizing the optimization process. We hope the result points in a promising direction toward modeling humans in motion and, eventually, achieving fully photorealistic, free-viewpoint rendering of people from casual captures.

\textbf{Acknowledgement:} We thank Marquese Scott for generously allowing us to feature his inspiring videos in this work. Special thanks to dear Lulu Chu for her enduring support. This work was funded by the UW Reality Lab, Meta, Google, Futurewei, and Amazon.

\clearpage
\section*{Supplementary Material}

\appendix

\section{Derivation of Motion Bases}

We describe how we derive the rotation and translation, $\{\rotbasis_i$, $\transbasis_i\}$, to map from bone coordinates in observation space to coordinates in canonical space (Section 3 on ``skeletal motion'').

We define body pose $\bodypose = (\joints, \jangles)$, where $\joints = \{\joint_i\}$ includes $K$ joint locations and $\jangles = \{\jangle_i\}$ defines local joint rotations using axis-angle representations $\in \mathfrak{so}(3)$. Given a predefined canonical pose $\bodypose_c = (\joints^c, \jangles^c)$ and an observed pose $\bodypose = (\joints, \jangles)$, the observation-to-canonical transformation $\spacetfm$ of body part $k$ is:

\begin{equation}
\begin{aligned}
\spacetfm_{k}&(\bodypose_c, \bodypose) \\
    &= \prod_{i \in \tau(k)}
        \begin{bmatrix}
            \exp(\jangle_i^c) && \joint_i^c \\
            0 && 1
        \end{bmatrix}
    \Bigg\{\prod_{i \in \tau(k)}
        \begin{bmatrix}
            \exp(\jangle_i) && \joint_i \\
            0 && 1
        \end{bmatrix}
    \Bigg\}^{-1},
\end{aligned}
\end{equation}
where $\exp(\jangle) \in SO(3)$ is a $3 \times 3$ rotation matrix computed by taking the exponential of $\jangle$ (i.e., applying Rodrigues' rotation formula), and $\tau(k)$ is the ordered set of parents of joint $K$ in the kinematic tree. 

The rotation and translation, $R_k$ and $t_k$, for body part $k$ is can then be extracted from $\spacetfm_{k}$:
\begin{equation}
    \begin{bmatrix}
            R_k && \mathbf{t}_k \\
            0 && 1
    \end{bmatrix}= 
    \spacetfm_{k}(\bodypose_c, \bodypose).
\end{equation}

\section{Network Architecture}

Figures 9-12 show the network design for the canonical MLP, the non-rigid motion MLP, the pose correction MLP, and the deep network generating the canonical motion weight volume.

\begin{figure}[htbp]
  \vspace{10px}
  \centering
  \includegraphics[width=\linewidth]{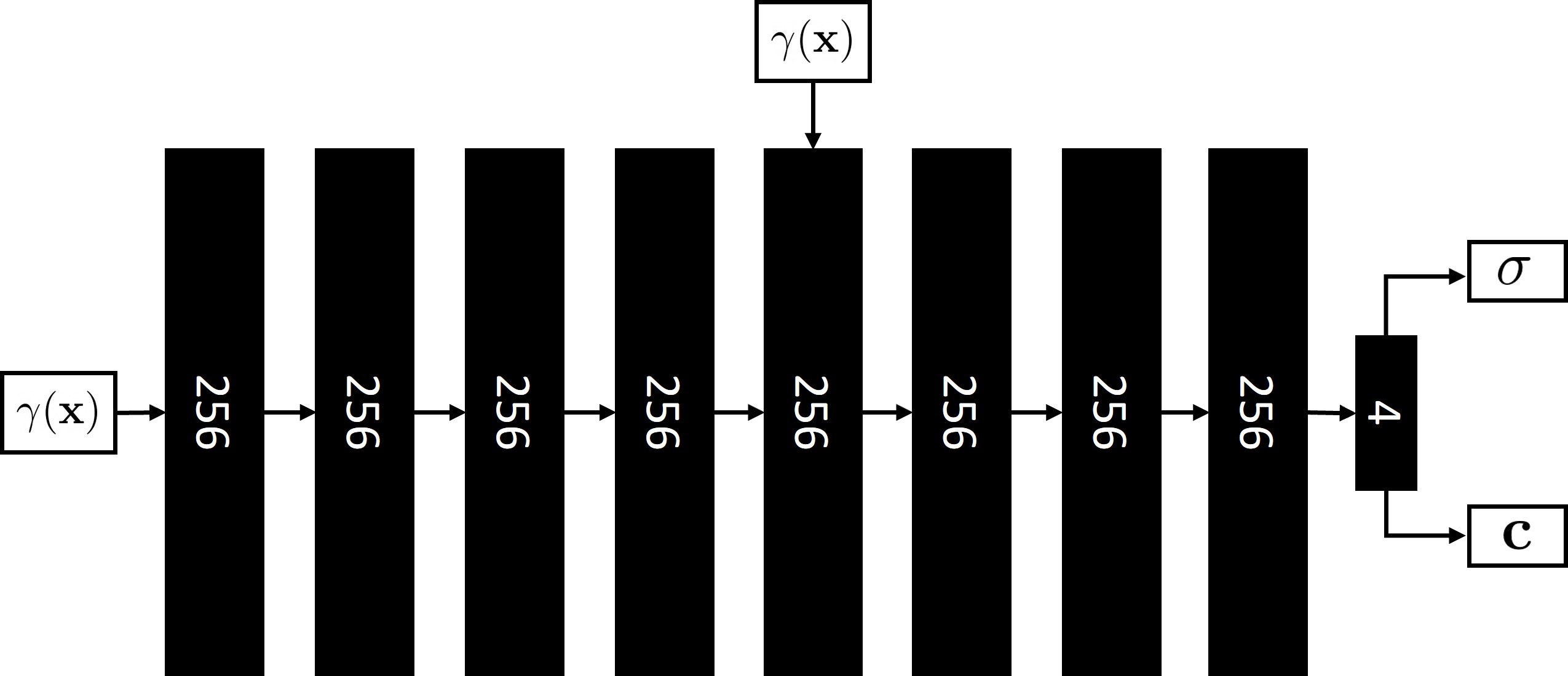}
  \caption{Canonical $\mlp$ visualization. Following NeRF \cite{mildenhall2020nerf}, we use an 8-layer MLP with width=256, taking as input positional encoding $\posencode$ of position $\pt$ and producing color $\mlpcolor$ and density $\mlpdensity$. A skip connection that concatenates $\posencode(\pt)$ to the fifth layer is applied. We adopt ReLU activation after each fully connected layer, except for the one generating color $\mlpcolor$ where we use \textit{sigmoid}.}
  \label{fig:mlp_canonical}
  \vspace{10px}
\end{figure}

\begin{figure}[htbp]
  \centering
  \includegraphics[width=0.75\linewidth]{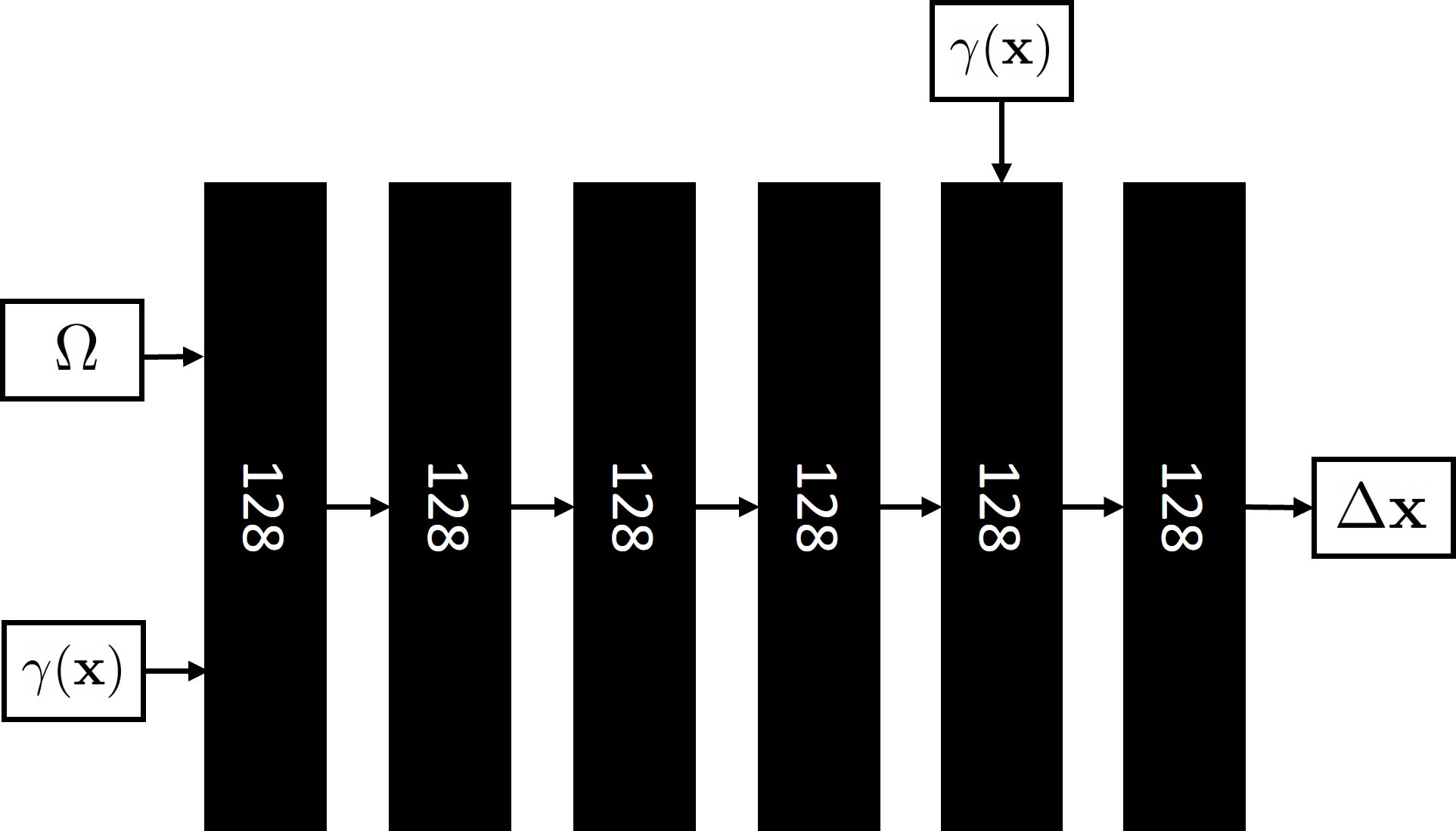}
  \caption{Non-rigid motion $\mlp$ visualization. We choose a 6-layer MLP (width=128) that takes as input the body pose, specifically, joint rotations $\jangles$, and positional encoding, $\posencode(\pt)$, and predicts the offset $\nroffsetpack$. We use a skip connection for the positional encoding at the fifth layer. Additionally, we remove the rotation vector of global orientation from joint angles $\jangles$ and only uses the remainder as MLP input.}
  \label{fig:mlp_non_rigid}
\end{figure}

\begin{figure}[htbp]
  \centering
  \includegraphics[width=0.6\linewidth]{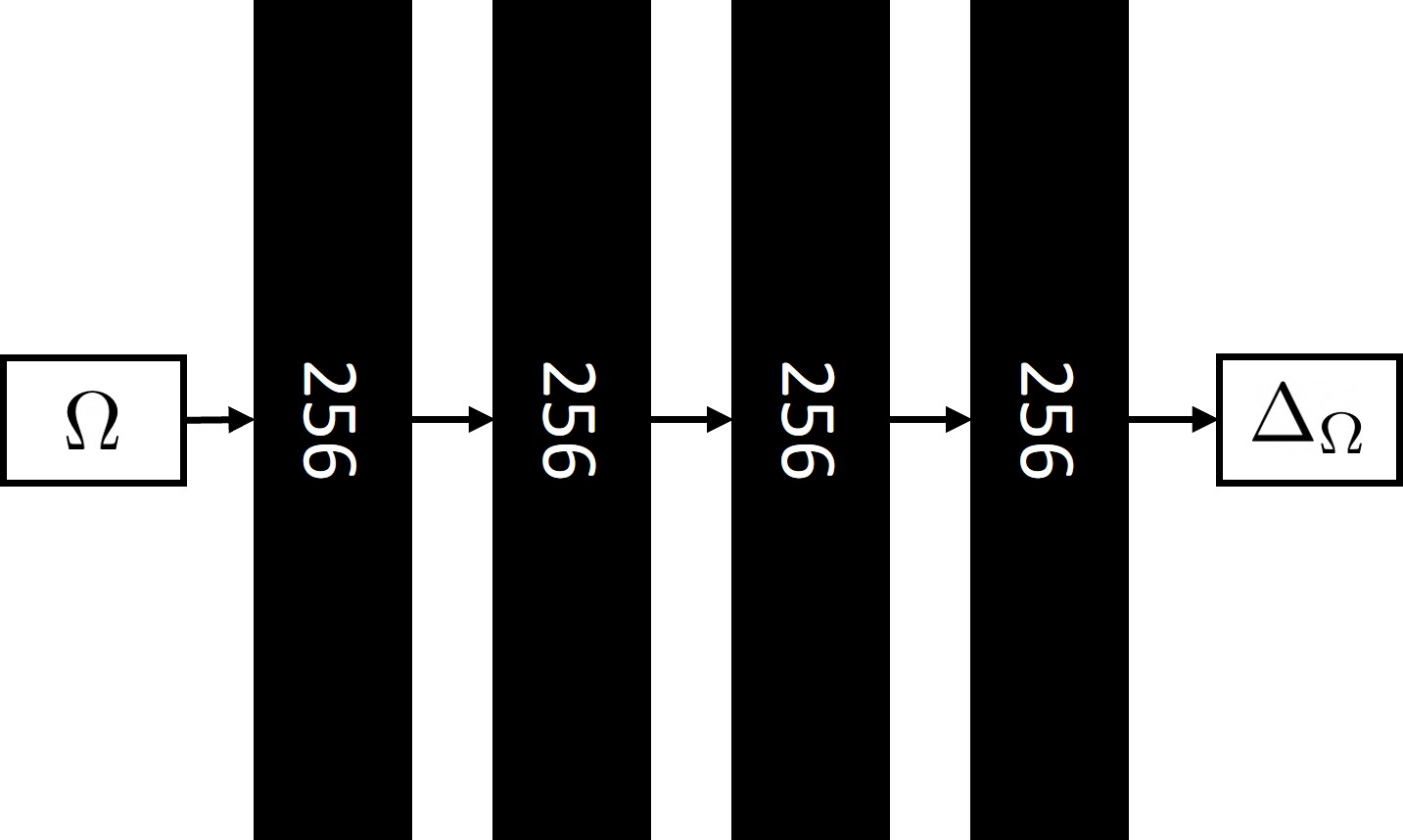}
  \caption{Pose correction $\mlp$ visualization. A 4-layer MLP with width 256 that takes joint angles $\jangles$ is used for refining initial poses. Like the non-rigid motion MLP, we take all joints except for root joint (i.e., body orientation) into account and optimize them accordingly.}
  \label{fig:mlp_pose_correct}
\end{figure}

\begin{figure}[htbp]
  \centering
  \includegraphics[width=\linewidth]{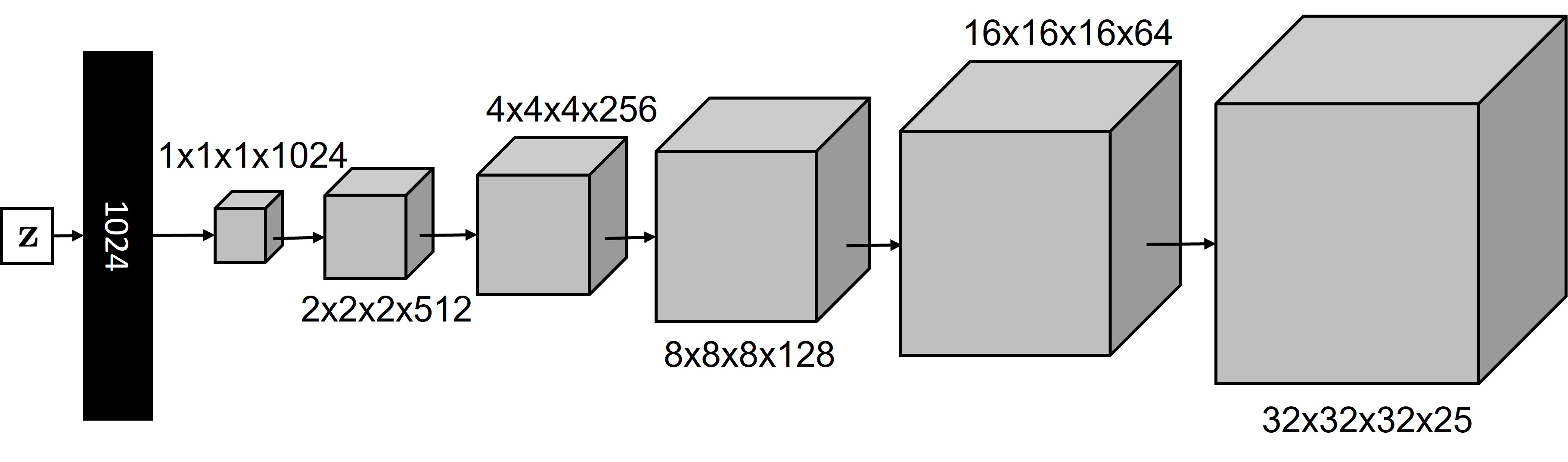}
  \caption{Network for generating the motion weight volume. The network begins with a fully-connected layer that transforms the (random, constant) latent code $\textbf{z}$ and reshapes it to a $1 \times 1 \times 1 \times 1024$ grid. Subsequently, it is concatenated with 5 transposed convolutions, increasing volume size while decreasing the number of channels, and finally, produces a volume of size $32 \times 32 \times 32 \times 25$.  LeakyReLU is applied after MLP and transposed convolution layers. The size of the latent code $\textbf{z}$ is 256. }
  \label{fig:network_motion_weight_volume}
\end{figure}

\section{Motion Field Decomposition}

\begin{figure*}[t]
  \centering
  \includegraphics[width=\linewidth]{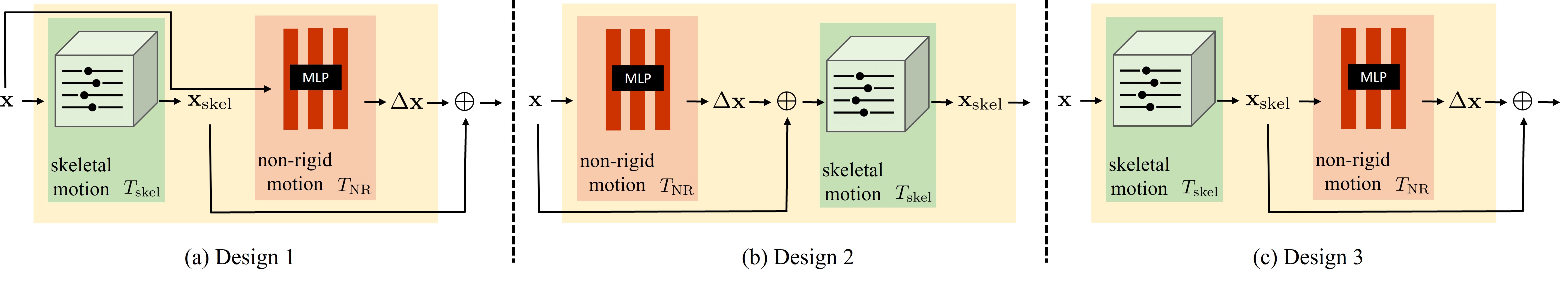}
  \caption{The three proposed designs of motion decomposition. We choose design 3 (c) as a result of best quality of novel view synthesis, shown in Fig.~\ref{fig:motion_decomposition_nvs}.}
  \label{fig:motion_decomposition_design}
\end{figure*}

We decompose a motion field into skeletal rigid motion and non-rigid motion. We tested several different formulations for the decomposition.  Specifically, starting from a point $\pt$ in observation space, we considered three potential decompositions. (To simplify notation and improve readability below, we omit body pose $\bodypose$, which would otherwise always appear as the second argument to each of $\motionfield, \skelmotionfield, \nrmotionfield$.)

(1) Both $\skelmotionfield$ and $\nrmotionfield$ conditioned on an observed point position $\pt$, illustrated in Fig.~\ref{fig:motion_decomposition_design}-(a):

\begin{equation}
\motionfield(\pt) = \skelmotionfield(\pt) 
  + \nrmotionfield(\pt)
\end{equation}

(2) $\nrmotionfield$ conditioned on $\pt$, but $\skelmotionfield$ conditioned on position adjusted by non-rigid motion, $\pt + \nrmotionfield(\pt)$, illustrated in Fig.~\ref{fig:motion_decomposition_design}-(b):

\begin{equation}
\motionfield(\pt) = \skelmotionfield(\pt + \nrmotionfield(\pt))
\end{equation}

(3) $\skelmotionfield$ conditioned on $\pt$ and $\nrmotionfield$ conditioned on the position $\skelmotionfield(\pt)$ warped by skeletal rigid motion $\skelmotionfield$, illustrated in Fig.~\ref{fig:motion_decomposition_design}-(c):

\begin{equation}
\motionfield(\pt) = \skelmotionfield(\pt) 
  + \nrmotionfield(\skelmotionfield(\pt))
\label{eq:skeletal_then_non_rigid}
\end{equation}

\begin{figure}[htbp]
  \centering
  \includegraphics[width=\linewidth]{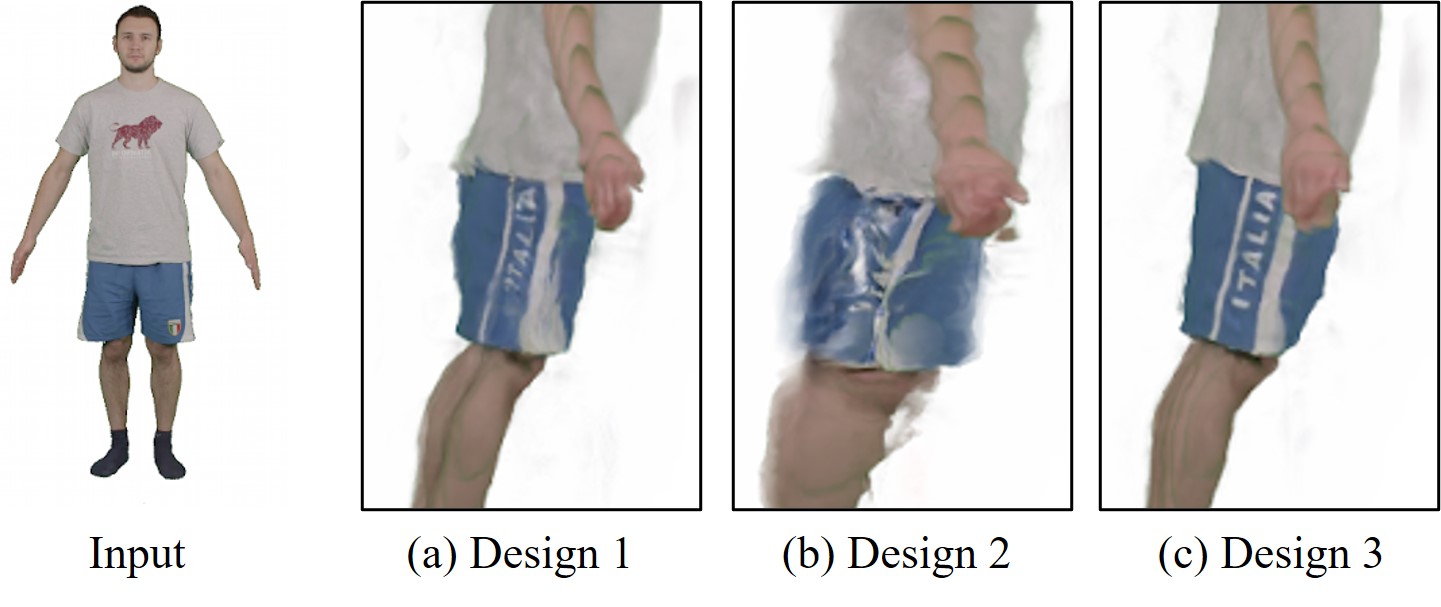}
  \caption{The experimental result of novel view synthesis on the three proposed motion decompositions, illustrated in Fig.~\ref{fig:motion_decomposition_design}. Design 3 (c) leads to best alignment, the approach we ultimately adopted. In this experiment, we used 64 samples per ray for quick evaluation, introducing color artifacts on the arms not present when using the sampling described in the paper.}
  \label{fig:motion_decomposition_nvs}
\end{figure}

We conducted experiments on the PeopleSnapshot dataset \cite{alldieck2018video}, and used 64 samples per ray for quick evaluation. As shown in Fig.~\ref{fig:motion_decomposition_nvs}, deforming $\pt$ by $\skelmotionfield$ and then conditioning $\nrmotionfield$ on that motion (design 3, or Eq.~\ref{eq:skeletal_then_non_rigid}) produces the best quality for novel view synthesis. The result of this experiment explains our final choice of motion decomposition.

\section{Additional Implementation Details}

There are several small but important implementation details that contribute to best results. We describe them below. 

\textbf{Optimizing $\mathbf{\Delta \weightvolcnl}$}: Our method solves for $\weightvolcnl$ to determine skeletal rigid motion. In practice, we ask a deep network to generate $\wightvoldelta$ instead, the difference between $\weightvolcnl$ and the logarithm of $\weightvolgaussian$. $\weightvolgaussian$ consists of an ellipsoidal Gaussian around each body bone, given by the canonical T-pose, that specifies approximate body part regions in the canonical space. $\weightvolcnl$ is then computed as:

\begin{equation}
    \weightvolcnl = \mathit{softmax}(\wightvoldelta + \log(\weightvolgaussian)), 
\end{equation}
where the background weight in $\weightvolgaussian$ is set to one minus the sum of all the bone weights. We apply the logarithm to $\weightvolgaussian$, to compensate the exponential function in \textit{softmax}.

\textbf{Representation of global body orientation:} Global subject orientation can be represented as body rotation or, equivalently, camera rotation. We choose to rotate the camera in order to keep the estimated bounding box when subject orientation changes.  Specifically, we uses axis-aligned bounding boxes because for ease of implementation; however, the box will be different for the same pose but rotated global body orientation. This undesirable effect can be avoided if we instead describe changes of global body orientation as camera rotations.

\textbf{Random background:} During optimization, we randomly assign a solid background color to the rendering and to the input image to facilitate separation of foreground and background. 

\renewcommand{\arraystretch}{1.2}
\begin{table*}[ht]
\centering
\begin{tabular}{|c || c | c | c || c | c | c || c | c | c|}
\hline
\multirow{2}{*}{} &  \multicolumn{3}{c||}{Subject \textbf{313}} & \multicolumn{3}{c||}{Subject \textbf{315}} & \multicolumn{3}{c|}{Subject \textbf{390}} \\ 
\cline{2-10}
 & PSNR $\uparrow$ & SSIM $\uparrow$ & LPIPS* $\downarrow$ & PSNR $\uparrow$ & SSIM $\uparrow$ & LPIPS* $\downarrow $ & PSNR $\uparrow$ & SSIM $\uparrow$ & LPIPS* $\downarrow$ \\ 
\hline
Neural Body \cite{peng2021neural} & 29.417 & 0.9635 & 57.24 & \cellcolor{best_color}26.93 & 0.9597 & 55.97 & 29.57 & 0.9609 & 52.12 \\
\hline
Ours & \cellcolor{best_color}29.421  & \cellcolor{best_color}0.9672 & \cellcolor{best_color}29.54 & 26.65 & \cellcolor{best_color}0.9636 & \cellcolor{best_color}33.76 & \cellcolor{best_color}30.52 & \cellcolor{best_color}0.9682 & \cellcolor{best_color}33.88 \\
\hline
\end{tabular}
\caption{Additional quantitative comparison on ZJU-MoCap dataset. We color cells having the \colorbox{best_color}{best} metric value. LPIPS* = LPIPS $\times 10^3$.}
\label{table:zju_number_vs_nb_additional}
\end{table*}

\textbf{MLP initialization}: We initialize the weights of the last layer of the non-rigid motion MLP and pose correction MLP to small values, $\mathcal{U}(-10^{-5}, 10^{-5})$, i.e., initializing the offset to be close to zero and the pose refinement rotation matrices each near the identity.

\textbf{Importance ray sampling}: We sample more rays for the foreground subject, indicated by the segmentation masks. Specifically, we enforce random ray sampling with probability 0.8 for foreground subject pixels and 0.2 for the background region.


\section{More Results}

\subsection{Additional Results}

We conduct an additional experiment on the remaining three subjects (313, 315, 390) in ZJU-MoCap dataset. The results are shown in Table~\ref{table:zju_number_vs_nb_additional}. Consistent with the results in the main paper, our method outperforms NeuralBody, particularly under the perceptual metric LPIPS. Fig.~\ref{fig:zju_mocap_vs_nb_more} shows  visual comparisons. Our method substantially captures the appearance details for unseen regions while Neural Body produces blurry results.

\subsection{Ablation Study on Sequence Length}

To understand how our method performs on different sequence lengths, we evaluate it on the sequences that vary in the number of frames but are sampled from the same video. Specifically, we take subject 392 from ZJU-MoCap dataset and use images captured from ``camera 1'' temporally sub-sampled at rates of 1, 2, 5, 10, and 30, yielding five training sequences containing 556, 228, 112, 56, and 19 frames respectively. For evaluation, we use the same motion sequence temporally sub-sampled by 30 but captured from the other 22 cameras not seen in the training. We use the same hyperparameters and training iterations throughout. The results are shown in Table~\ref{table:ablation_seq_length}.

\begin{table}[H]
\centering
\begin{tabular}{|c|c|c|c|}
\hline
 & PSNR $\uparrow$ & SSIM $\uparrow$ & LPIPS* $\downarrow$ \\
\hline 
\hline
556 frames & \cellcolor{best_color}31.04 & \cellcolor{best_color}0.9705 & 32.12 \\
\hline
228 frames & 30.84 & 0.9701 &  \cellcolor{best_color}31.78 \\
\hline
112 frames & 31.01 & 0.9703 & 32.75 \\
\hline
56 frames & 30.90 & 0.9693 &  35.45 \\
\hline
19 frames & 30.51 & 0.9655 &  45.17 \\
\hline
\end{tabular}
\caption{Ablation study on sequence length. We color cells with \colorbox{best_color}{best} metric values. LPIPS* = LPIPS $\times 10^3$.}
\label{table:ablation_seq_length}
\end{table}

As expected, using more frames leads to better quality; however the improvement is not obvious when the frame number is over a threshold (in this case, 112 frames). We speculate that diversity of body poses is a more significant factor in reconstruction quality than number of frames.

\subsection{Optimized Canonical Appearance}

Fig.~\ref{fig:zju_tpose} shows the recovered appearance for the pre-defined T-pose on the ZJU-MoCap ~\cite{peng2021neural} dataset; the results for self-captured and YouTube videos are shown in Fig.~\ref{fig:wild_tpose}.

\subsection{Limitations}

\begin{figure}[htbp]
  \centering
  \includegraphics[width=\linewidth]{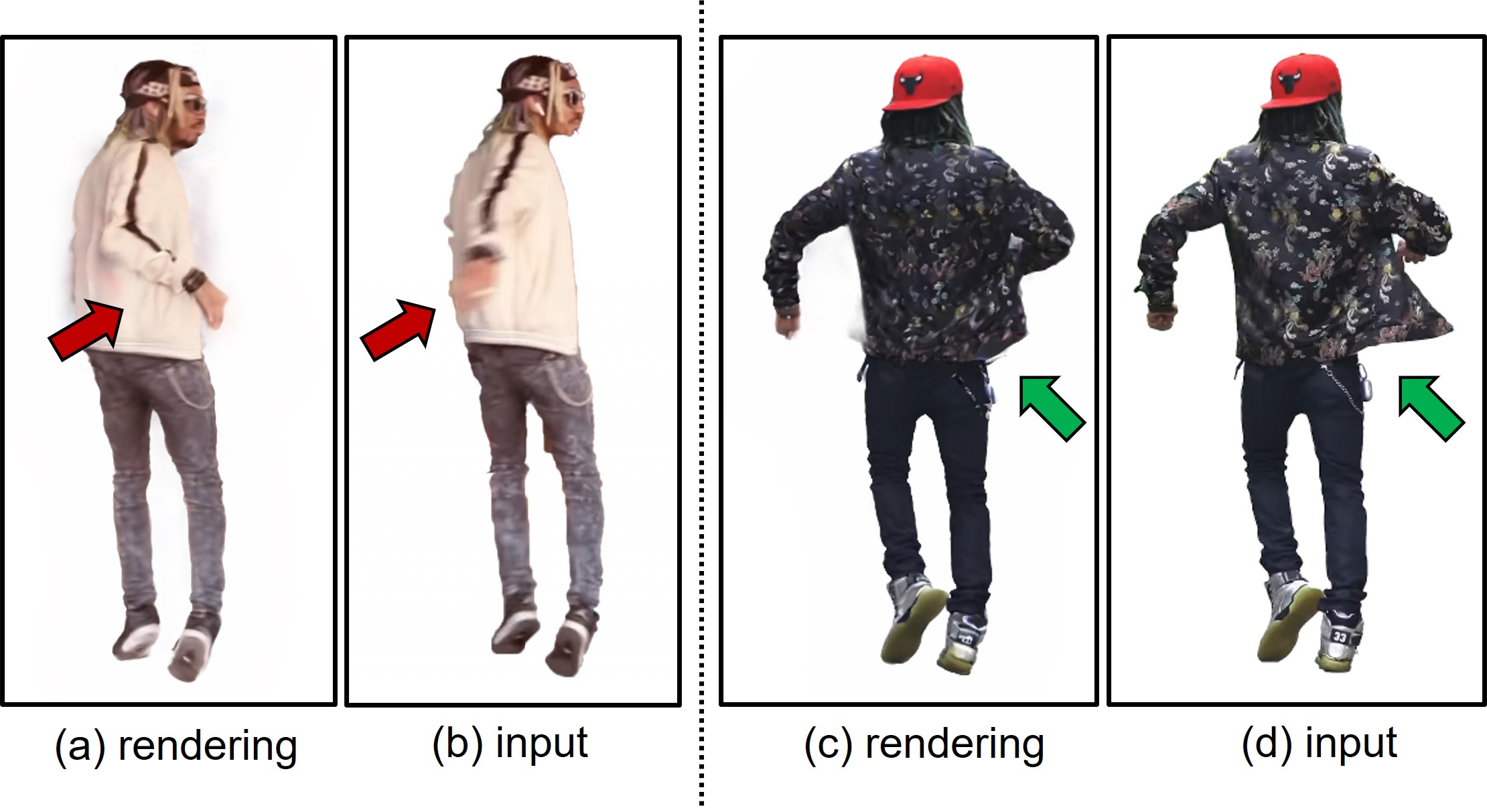}
  \caption{Visual examples of limitations. Pose correction may fail (a and b) and non-rigid clothes motion was not able to be fully recovered (c and d).}
  \vspace{-10px}
  \label{fig:limitation}
\end{figure}

\begin{figure*}
  \centering
  \includegraphics[width=0.95\textwidth]{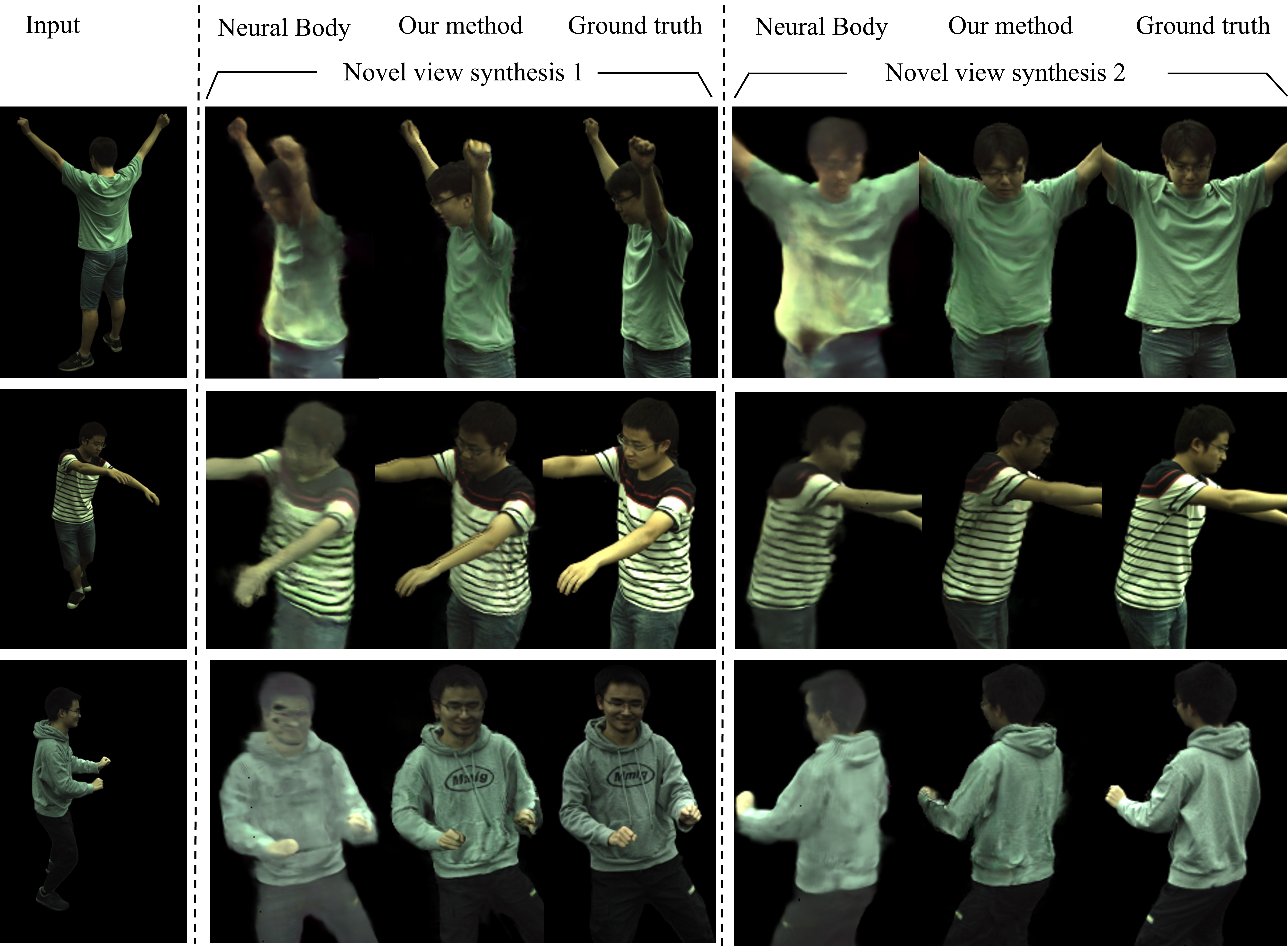}
  \caption{Qualitative comparison on the remaining subjects in ZJU-MoCap dataset.}
  \vspace{-10px}
  \label{fig:zju_mocap_vs_nb_more}
\end{figure*}

\begin{figure*}[htbp]
  \centering
  \includegraphics[width=\textwidth]{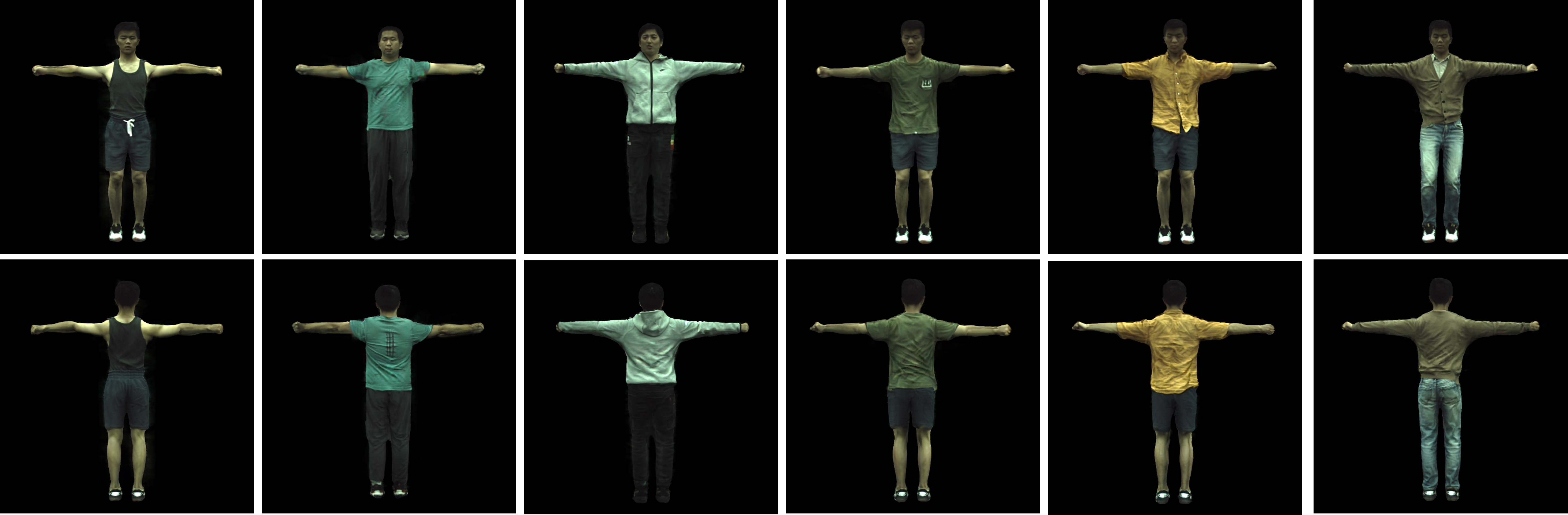}
  \caption{Optimized canonical appearance on  ZJU-MoCap dataset.}
  \label{fig:zju_tpose}
\end{figure*}

\begin{figure*}[htbp]
  \centering
  \includegraphics[width=\textwidth]{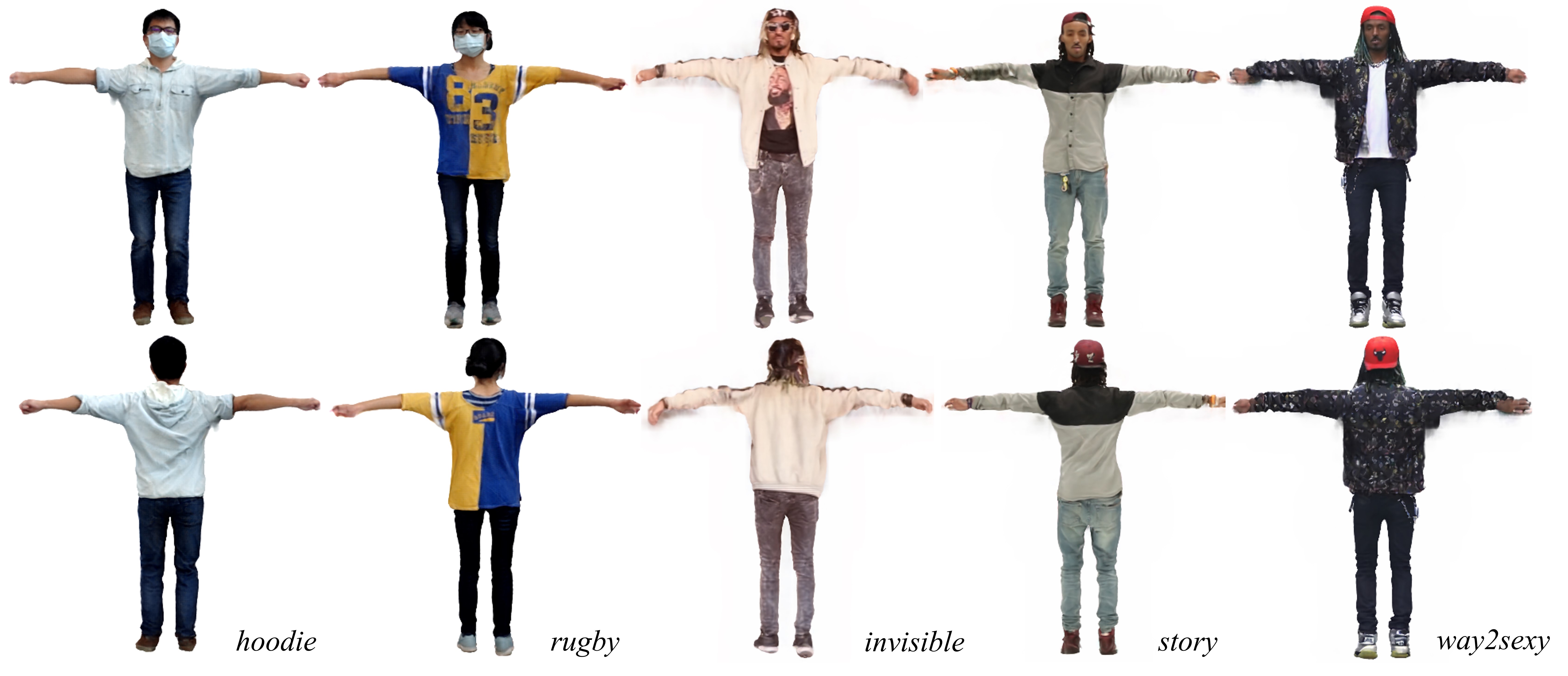}
  \caption{Optimized canonical appearance for self-captured videos (first two columns) and YouTube videos (right three).}
  \label{fig:wild_tpose}
\end{figure*}

We provide two visual examples of our method's limitations in Fig.~\ref{fig:limitation}. Pose correction may fail if the video frame contains artifacts, e.g., strong motion blur, as shown in (a) and (b).  Non-rigid motion was not fully recovered in (c) and (d), as the movement of the jacket depended on the temporal dynamics of subject motion.

\section{Societal Impact}

In this work we aim to faithfully reproduce motion sequences performed by a person with the capability of rendering unseen views. Therefore applying the technology to create false depictions, e.g., re-animating the subject in novel poses, was not considered as a potential application. Nevertheless, the public deployment of the technology should still be done with care, e.g., by reminding audiences that imagery is computer-generated when adjusting the viewpoint. In addition, the high computation requirement of the algorithm may lead to increased carbon emissions. We hope the methods that accelerate training of neural graphics primitives (e.g., \cite{mueller2022instant}) will help reduce computation and thus the environmental impact. Finally, our method will be made available to the public for counter-measure analysis and computation reduction.

{\small
\bibliographystyle{ieee_fullname}
\bibliography{egbib}

\begin{thebibliography}{10}\itemsep=-1pt

\bibitem{alldieck2018video}
Thiemo Alldieck, Marcus Magnor, Weipeng Xu, Christian Theobalt, and Gerard
  Pons-Moll.
\newblock Video based reconstruction of 3d people models.
\newblock In {\em {IEEE} Conference on Computer Vision and Pattern
  Recognition}, 2018.

\bibitem{balakrishnan2018synthesizing}
Guha Balakrishnan, Amy Zhao, Adrian~V Dalca, Fredo Durand, and John Guttag.
\newblock Synthesizing images of humans in unseen poses.
\newblock {\em CVPR}, 2018.

\bibitem{barron2021mipnerf}
Jonathan~T. Barron, Ben Mildenhall, Matthew Tancik, Peter Hedman, Ricardo
  Martin-Brualla, and Pratul~P. Srinivasan.
\newblock {Mip-NeRF}: A multiscale representation for anti-aliasing neural
  radiance fields.
\newblock {\em ICCV}, 2021.

\bibitem{bhatnagar2020loopreg}
Bharat~Lal Bhatnagar, Cristian Sminchisescu, Christian Theobalt, and Gerard
  Pons-Moll.
\newblock Loopreg: Self-supervised learning of implicit surface
  correspondences, pose and shape for 3d human mesh registration.
\newblock {\em Advances in Neural Information Processing Systems}, 33, 2020.

\bibitem{carranza2003free}
Joel Carranza, Christian Theobalt, Marcus~A Magnor, and Hans-Peter Seidel.
\newblock Free-viewpoint video of human actors.
\newblock {\em ACM transactions on graphics (TOG)}, 2003.

\bibitem{casas20144d}
Dan Casas, Marco Volino, John Collomosse, and Adrian Hilton.
\newblock {4D} video textures for interactive character appearance.
\newblock {\em Computer Graphics Forum}, 2014.

\bibitem{chan2019everybody}
Caroline Chan, Shiry Ginosar, Tinghui Zhou, and Alexei~A Efros.
\newblock Everybody dance now.
\newblock {\em ICCV}, 2019.

\bibitem{chaurasia2013depth}
Gaurav Chaurasia, Sylvain Duchene, Olga Sorkine-Hornung, and George Drettakis.
\newblock Depth synthesis and local warps for plausible image-based navigation.
\newblock {\em ACM Transactions on Graphics (TOG)}, 2013.

\bibitem{chen1993view}
Shenchang~Eric Chen and Lance Williams.
\newblock View interpolation for image synthesis.
\newblock {\em SIGGRAPH}, 1993.

\bibitem{chen2021snarf}
Xu Chen, Yufeng Zheng, Michael~J Black, Otmar Hilliges, and Andreas Geiger.
\newblock {SNARF}: Differentiable forward skinning for animating non-rigid
  neural implicit shapes.
\newblock {\em ICCV}, 2021.

\bibitem{collet2015high}
Alvaro Collet, Ming Chuang, Pat Sweeney, Don Gillett, Dennis Evseev, David
  Calabrese, Hugues Hoppe, Adam Kirk, and Steve Sullivan.
\newblock High-quality streamable free-viewpoint video.
\newblock {\em ACM Transactions on Graphics (ToG)}, 2015.

\bibitem{de2008performance}
Edilson De~Aguiar, Carsten Stoll, Christian Theobalt, Naveed Ahmed, Hans-Peter
  Seidel, and Sebastian Thrun.
\newblock Performance capture from sparse multi-view video.
\newblock {\em SIGGRAPH}, 2008.

\bibitem{debevec1996modeling}
Paul~E Debevec, Camillo~J Taylor, and Jitendra Malik.
\newblock Modeling and rendering architecture from photographs: A hybrid
  geometry-and image-based approach.
\newblock {\em SIGGRAPH}, 1996.

\bibitem{deng2019neural}
Boyang Deng, JP Lewis, Timothy Jeruzalski, Gerard Pons-Moll, Geoffrey Hinton,
  Mohammad Norouzi, and Andrea Tagliasacchi.
\newblock Neural articulated shape approximation.
\newblock {\em ECCV}, 2020.

\bibitem{flagg2009human}
Matthew Flagg, Atsushi Nakazawa, Qiushuang Zhang, Sing~Bing Kang, Young~Kee
  Ryu, Irfan Essa, and James~M Rehg.
\newblock Human video textures.
\newblock In {\em Proceedings of the 2009 symposium on Interactive 3D graphics
  and games}, pages 199--206, 2009.

\bibitem{Gao-freeviewvideo}
Chen Gao, Ayush Saraf, Johannes Kopf, and Jia-Bin Huang.
\newblock Dynamic view synthesis from dynamic monocular video.
\newblock {\em ICCV}, 2021.

\bibitem{gortler1996lumigraph}
Steven~J Gortler, Radek Grzeszczuk, Richard Szeliski, and Michael~F Cohen.
\newblock The lumigraph.
\newblock {\em SIGGRAPH}, 1996.

\bibitem{guo2019relightables}
Kaiwen Guo, Peter Lincoln, Philip Davidson, Jay Busch, Xueming Yu, Matt Whalen,
  Geoff Harvey, Sergio Orts-Escolano, Rohit Pandey, Jason Dourgarian, et~al.
\newblock The relightables: Volumetric performance capture of humans with
  realistic relighting.
\newblock {\em ACM Transactions on Graphics (TOG)}, 2019.

\bibitem{habermann2021}
Marc Habermann, Lingjie Liu, Weipeng Xu, Michael Zollhoefer, Gerard Pons-Moll,
  and Christian Theobalt.
\newblock Real-time deep dynamic characters.
\newblock {\em ACM Transactions on Graphics (TOG)}, 2021.

\bibitem{he2021arch++}
Tong He, Yuanlu Xu, Shunsuke Saito, Stefano Soatto, and Tony Tung.
\newblock Arch++: Animation-ready clothed human reconstruction revisited.
\newblock In {\em Proceedings of the IEEE/CVF International Conference on
  Computer Vision}, pages 11046--11056, 2021.

\bibitem{hedman2018instant}
Peter Hedman and Johannes Kopf.
\newblock Instant {3D} photography.
\newblock {\em ACM Transactions on Graphics (TOG)}, 37(4):1--12, 2018.

\bibitem{hedman2016scalable}
Peter Hedman, Tobias Ritschel, George Drettakis, and Gabriel Brostow.
\newblock Scalable inside-out image-based rendering.
\newblock {\em ACM Transactions on Graphics (TOG)}, 2016.

\bibitem{hedman2021baking}
Peter Hedman, Pratul~P. Srinivasan, Ben Mildenhall, Jonathan~T. Barron, and
  Paul Debevec.
\newblock Baking neural radiance fields for real-time view synthesis.
\newblock {\em ICCV}, 2021.

\bibitem{hertz2021sape}
Amir Hertz, Or Perel, Raja Giryes, Olga Sorkine-Hornung, and Daniel Cohen-Or.
\newblock {SAPE}: Spatially-adaptive progressive encoding for neural
  optimization.
\newblock {\em arXiv:2104.09125}, 2021.

\bibitem{huang2020arch}
Zeng Huang, Yuanlu Xu, Christoph Lassner, Hao Li, and Tony Tung.
\newblock Arch: Animatable reconstruction of clothed humans.
\newblock {\em CVPR}, 2020.

\bibitem{zhang2021stnerf}
Zhang Jiakai, Liu Xinhang, Ye Xinyi, Zhao Fuqiang, Zhang Yanshun, Wu Minye,
  Zhang Yingliang, Xu Lan, and Yu Jingyi.
\newblock Editable free-viewpoint video using a layered neural representation.
\newblock {\em ACM SIGGRAPH}, 2021.

\bibitem{kanade1997virtualized}
Takeo Kanade, Peter Rander, and PJ Narayanan.
\newblock Virtualized reality: Constructing virtual worlds from real scenes.
\newblock {\em IEEE multimedia}, 1997.

\bibitem{kingma2014adam}
Diederik~P Kingma and Jimmy Ba.
\newblock Adam: A method for stochastic optimization.
\newblock {\em ICLR}, 2015.

\bibitem{kolotouros2019spin}
Nikos Kolotouros, Georgios Pavlakos, Michael~J. Black, and Kostas Daniilidis.
\newblock Learning to reconstruct {3D} human pose and shape via model-fitting
  in the loop.
\newblock {\em ICCV}, 2019.

\bibitem{levoy1996light}
Marc Levoy and Pat Hanrahan.
\newblock Light field rendering.
\newblock {\em SIGGRAPH}, 1996.

\bibitem{li2021neural}
Zhengqi Li, Simon Niklaus, Noah Snavely, and Oliver Wang.
\newblock Neural scene flow fields for space-time view synthesis of dynamic
  scenes.
\newblock {\em CVPR}, 2021.

\bibitem{liu2021neural}
Lingjie Liu, Marc Habermann, Viktor Rudnev, Kripasindhu Sarkar, Jiatao Gu, and
  Christian Theobalt.
\newblock Neural actor: Neural free-view synthesis of human actors with pose
  control.
\newblock {\em ACM Trans. Graph.(ACM SIGGRAPH Asia)}, 2021.

\bibitem{liu2020NeuralHumanRendering}
Lingjie Liu, Weipeng Xu, Marc Habermann, Michael Zollhöfer, Florian Bernard,
  Hyeongwoo Kim, Wenping Wang, and Christian Theobalt.
\newblock Neural human video rendering by learning dynamic textures and
  rendering-to-video translation.
\newblock {\em IEEE Transactions on Visualization and Computer Graphics}, 2020.

\bibitem{loper2015smpl}
Matthew Loper, Naureen Mahmood, Javier Romero, Gerard Pons-Moll, and Michael~J
  Black.
\newblock {SMPL}: A skinned multi-person linear model.
\newblock {\em ACM transactions on graphics (TOG)}, 2015.

\bibitem{ma2017pose}
Liqian Ma, Xu Jia, Qianru Sun, Bernt Schiele, Tinne Tuytelaars, and Luc
  Van~Gool.
\newblock Pose guided person image generation.
\newblock {\em arXiv:1705.09368}, 2017.

\bibitem{martin2018lookingood}
Ricardo Martin-Brualla, Rohit Pandey, Shuoran Yang, Pavel Pidlypenskyi,
  Jonathan Taylor, Julien Valentin, Sameh Khamis, Philip Davidson, Anastasia
  Tkach, Peter Lincoln, et~al.
\newblock {LookinGood}: Enhancing performance capture with real-time neural
  re-rendering.
\newblock {\em ACM Transactions on Graphics (TOG)}, 2018.

\bibitem{martinbrualla2020nerfw}
Ricardo Martin-Brualla, Noha Radwan, Mehdi S.~M. Sajjadi, Jonathan~T. Barron,
  Alexey Dosovitskiy, and Daniel Duckworth.
\newblock {NeRF} in the wild: Neural radiance fields for unconstrained photo
  collections.
\newblock {\em CVPR}, 2021.

\bibitem{matusik2000image}
Wojciech Matusik, Chris Buehler, Ramesh Raskar, Steven~J Gortler, and Leonard
  McMillan.
\newblock Image-based visual hulls.
\newblock {\em SIGGRAPH}, 2000.

\bibitem{max1995optical}
Nelson Max.
\newblock Optical models for direct volume rendering.
\newblock {\em IEEE Transactions on Visualization and Computer Graphics}, 1995.

\bibitem{LEAP:CVPR:21}
Marko Mihajlovic, Yan Zhang, Michael~J Black, and Siyu Tang.
\newblock {LEAP}: Learning articulated occupancy of people.
\newblock {\em CVPR}, 2021.

\bibitem{mildenhall2020nerf}
Ben Mildenhall, Pratul~P. Srinivasan, Matthew Tancik, Jonathan~T. Barron, Ravi
  Ramamoorthi, and Ren Ng.
\newblock {NeRF}: Representing scenes as neural radiance fields for view
  synthesis.
\newblock {\em ECCV}, 2020.

\bibitem{mueller2022instant}
Thomas M\"uller, Alex Evans, Christoph Schied, and Alexander Keller.
\newblock Instant neural graphics primitives with a multiresolution hash
  encoding.
\newblock {\em arXiv:2201.05989}, Jan. 2022.

\bibitem{neverova2018dense}
Natalia Neverova, Riza~Alp Guler, and Iasonas Kokkinos.
\newblock Dense pose transfer.
\newblock {\em ECCV}, 2018.

\bibitem{niemeyer2021giraffe}
Michael Niemeyer and Andreas Geiger.
\newblock {GIRAFFE}: Representing scenes as compositional generative neural
  feature fields.
\newblock {\em CVPR}, 2021.

\bibitem{2021narf}
Atsuhiro Noguchi, Xiao Sun, Stephen Lin, and Tatsuya Harada.
\newblock Neural articulated radiance field.
\newblock {\em ICCV}, 2021.

\bibitem{pandey2019volumetric}
Rohit Pandey, Anastasia Tkach, Shuoran Yang, Pavel Pidlypenskyi, Jonathan
  Taylor, Ricardo Martin-Brualla, Andrea Tagliasacchi, George Papandreou,
  Philip Davidson, Cem Keskin, et~al.
\newblock Volumetric capture of humans with a single {RGBD} camera via
  semi-parametric learning.
\newblock {\em CVPR}, 2019.

\bibitem{park2021nerfies}
Keunhong Park, Utkarsh Sinha, Jonathan~T. Barron, Sofien Bouaziz, Dan~B
  Goldman, Steven~M. Seitz, and Ricardo Martin-Brualla.
\newblock Nerfies: Deformable neural radiance fields.
\newblock {\em ICCV}, 2021.

\bibitem{park2021hypernerf}
Keunhong Park, Utkarsh Sinha, Peter Hedman, Jonathan~T. Barron, Sofien Bouaziz,
  Dan~B Goldman, Ricardo Martin-Brualla, and Steven~M. Seitz.
\newblock {HyperNeRF}: A higher-dimensional representation for topologically
  varying neural radiance fields.
\newblock {\em SIGGRAPH Asia}, 2021.

\bibitem{peng2021animatable}
Sida Peng, Junting Dong, Qianqian Wang, Shangzhan Zhang, Qing Shuai, Xiaowei
  Zhou, and Hujun Bao.
\newblock Animatable neural radiance fields for modeling dynamic human bodies.
\newblock {\em ICCV}, 2021.

\bibitem{peng2021neural}
Sida Peng, Yuanqing Zhang, Yinghao Xu, Qianqian Wang, Qing Shuai, Hujun Bao,
  and Xiaowei Zhou.
\newblock Neural body: Implicit neural representations with structured latent
  codes for novel view synthesis of dynamic humans.
\newblock {\em CVPR}, 2021.

\bibitem{pumarola2020d}
Albert Pumarola, Enric Corona, Gerard Pons-Moll, and Francesc Moreno-Noguer.
\newblock {D-NeRF}: Neural radiance fields for dynamic scenes.
\newblock {\em CVPR}, 2020.

\bibitem{Saito:CVPR:2021}
Shunsuke Saito, Jinlong Yang, Qianli Ma, and Michael~J. Black.
\newblock {SCANimate}: Weakly supervised learning of skinned clothed avatar
  networks.
\newblock {\em CVPR}, 2021.

\bibitem{sanyal2021learning}
Soubhik Sanyal, Alex Vorobiov, Timo Bolkart, Matthew Loper, Betty Mohler,
  Larry~S Davis, Javier Romero, and Michael~J Black.
\newblock Learning realistic human reposing using cyclic self-supervision with
  {3D} shape, pose, and appearance consistency.
\newblock In {\em CVPR}, 2021.

\bibitem{sarkar2021style}
Kripasindhu Sarkar, Vladislav Golyanik, Lingjie Liu, and Christian Theobalt.
\newblock Style and pose control for image synthesis of humans from a single
  monocular view.
\newblock {\em arXiv preprint arXiv:2102.11263}, 2021.

\bibitem{schoenberger2016mvs}
Johannes~Lutz Sch\"{o}nberger, Enliang Zheng, Marc Pollefeys, and Jan-Michael
  Frahm.
\newblock Pixelwise view selection for unstructured multi-view stereo.
\newblock {\em ECCV}, 2016.

\bibitem{Schwarz2020NEURIPS}
Katja Schwarz, Yiyi Liao, Michael Niemeyer, and Andreas Geiger.
\newblock {GRAF}: Generative radiance fields for 3d-aware image synthesis.
\newblock {\em NeurIPS}, 2020.

\bibitem{shum2000imagerendering}
Harry Shum and Sing~Bing Kang.
\newblock {Review of image-based rendering techniques}.
\newblock {\em Visual Communications and Image Processing 2000}, 2000.

\bibitem{srinivasan2021nerv}
Pratul~P. Srinivasan, Boyang Deng, Xiuming Zhang, Matthew Tancik, Ben
  Mildenhall, and Jonathan~T. Barron.
\newblock {NeRV}: Neural reflectance and visibility fields for relighting and
  view synthesis.
\newblock {\em CVPR}, 2021.

\bibitem{starck2005video}
Jonathan Starck, Gregor Miller, and Adrian Hilton.
\newblock Video-based character animation.
\newblock {\em ACM SIGGRAPH/Eurographics symposium on Computer animation},
  2005.

\bibitem{su2021anerf}
Shih-Yang Su, Frank Yu, Michael Zollh{\"o}fer, and Helge Rhodin.
\newblock A-nerf: Articulated neural radiance fields for learning human shape,
  appearance, and pose.
\newblock In {\em Advances in Neural Information Processing Systems}, 2021.

\bibitem{szeliski2010computer}
Richard Szeliski.
\newblock {\em Computer vision: algorithms and applications}.
\newblock Springer Science \& Business Media, 2010.

\bibitem{tancik2020fourfeat}
Matthew Tancik, Pratul~P. Srinivasan, Ben Mildenhall, Sara Fridovich-Keil,
  Nithin Raghavan, Utkarsh Singhal, Ravi Ramamoorthi, Jonathan~T. Barron, and
  Ren Ng.
\newblock Fourier features let networks learn high frequency functions in low
  dimensional domains.
\newblock {\em NeurIPS}, 2020.

\bibitem{tiwari2021neural}
Garvita Tiwari, Nikolaos Sarafianos, Tony Tung, and Gerard Pons-Moll.
\newblock Neural-gif: Neural generalized implicit functions for animating
  people in clothing.
\newblock In {\em Proceedings of the IEEE/CVF International Conference on
  Computer Vision}, pages 11708--11718, 2021.

\bibitem{tretschk2021nonrigid}
Edgar Tretschk, Ayush Tewari, Vladislav Golyanik, Michael Zollh\"{o}fer,
  Christoph Lassner, and Christian Theobalt.
\newblock Non-rigid neural radiance fields: Reconstruction and novel view
  synthesis of a dynamic scene from monocular video.
\newblock {\em ICCV}, 2021.

\bibitem{vlasic2008articulated}
Daniel Vlasic, Ilya Baran, Wojciech Matusik, and Jovan Popovi{\'c}.
\newblock Articulated mesh animation from multi-view silhouettes.
\newblock {\em SIGGRAPH}, 2008.

\bibitem{wang2021ibrnet}
Qianqian Wang, Zhicheng Wang, Kyle Genova, Pratul Srinivasan, Howard Zhou,
  Jonathan~T. Barron, Ricardo Martin-Brualla, Noah Snavely, and Thomas
  Funkhouser.
\newblock {IBRNet}: Learning multi-view image-based rendering.
\newblock {\em CVPR}, 2021.

\bibitem{wang2018video}
Ting-Chun Wang, Ming-Yu Liu, Jun-Yan Zhu, Guilin Liu, Andrew Tao, Jan Kautz,
  and Bryan Catanzaro.
\newblock Video-to-video synthesis.
\newblock {\em NeurIPS}, 2018.

\bibitem{wang2021dance}
Tuanfeng~Y Wang, Duygu Ceylan, Krishna~Kumar Singh, and Niloy~J Mitra.
\newblock Dance in the wild: Monocular human animation with neural dynamic
  appearance synthesis.
\newblock In {\em 2021 International Conference on 3D Vision (3DV)}, pages
  268--277. IEEE, 2021.

\bibitem{weng2020vid2actor}
Chung-Yi Weng, Brian Curless, and Ira Kemelmacher-Shlizerman.
\newblock {Vid2Actor}: Free-viewpoint animatable person synthesis from video in
  the wild.
\newblock {\em arXiv:2012.12884}, 2020.

\bibitem{wu2020multi}
Minye Wu, Yuehao Wang, Qiang Hu, and Jingyi Yu.
\newblock Multi-view neural human rendering.
\newblock {\em CVPR}, 2020.

\bibitem{xian2021space}
Wenqi Xian, Jia-Bin Huang, Johannes Kopf, and Changil Kim.
\newblock Space-time neural irradiance fields for free-viewpoint video.
\newblock {\em CVPR}, 2021.

\bibitem{xu2011video}
Feng Xu, Yebin Liu, Carsten Stoll, James Tompkin, Gaurav Bharaj, Qionghai Dai,
  Hans-Peter Seidel, Jan Kautz, and Christian Theobalt.
\newblock Video-based characters: creating new human performances from a
  multi-view video database.
\newblock {\em SIGGRAPH}, 2011.

\bibitem{xu2021h}
Hongyi Xu, Thiemo Alldieck, and Cristian Sminchisescu.
\newblock H-nerf: Neural radiance fields for rendering and temporal
  reconstruction of humans in motion.
\newblock {\em Advances in Neural Information Processing Systems}, 34, 2021.

\bibitem{yang2021s3}
Ze Yang, Shenlong Wang, Siva Manivasagam, Zeng Huang, Wei-Chiu Ma, Xinchen Yan,
  Ersin Yumer, and Raquel Urtasun.
\newblock S3: Neural shape, skeleton, and skinning fields for 3d human
  modeling.
\newblock {\em CVPR}, 2021.

\bibitem{zhang2020nerf++}
Kai Zhang, Gernot Riegler, Noah Snavely, and Vladlen Koltun.
\newblock {NeRF++}: Analyzing and improving neural radiance fields.
\newblock {\em arXiv:2010.07492}, 2020.

\bibitem{zhang2018unreasonable}
Richard Zhang, Phillip Isola, Alexei~A Efros, Eli Shechtman, and Oliver Wang.
\newblock The unreasonable effectiveness of deep features as a perceptual
  metric.
\newblock {\em CVPR}, 2018.

\bibitem{nerfactor}
Xiuming Zhang, Pratul~P. Srinivasan, Boyang Deng, Paul Debevec, William~T.
  Freeman, and Jonathan~T. Barron.
\newblock {NeRFactor}: Neural factorization of shape and reflectance under an
  unknown illumination.
\newblock {\em SIGGRAPH Asia}, 2021.

\bibitem{zitnick2004high}
C~Lawrence Zitnick, Sing~Bing Kang, Matthew Uyttendaele, Simon Winder, and
  Richard Szeliski.
\newblock High-quality video view interpolation using a layered representation.
\newblock {\em ACM transactions on graphics (TOG)}, 2004.

\end{thebibliography}
}

\end{document}